\documentclass{article}

\usepackage{PRIMEarxiv}

\usepackage{diagbox}
\usepackage{lipsum}
\usepackage{natbib}
\usepackage{tabularx}
\setcitestyle{numbers,square}
\usepackage{multicol}
\usepackage{amssymb}
\usepackage{graphicx}%
\usepackage{multirow}%
\usepackage{amsmath,amssymb,amsfonts}%
\usepackage{amsthm}%
\usepackage{mathrsfs}%
\usepackage[title]{appendix}%
\usepackage{xcolor}%
\usepackage{textcomp}%
\usepackage{rotating}
\usepackage{manyfoot}%
\usepackage{booktabs}%
\usepackage[ruled,linesnumbered]{algorithm2e}
\usepackage{algpseudocode}%
\usepackage{listings}%
\usepackage[justification=centering]{caption}%
\usepackage{makecell}
\usepackage{CJKutf8}
\usepackage{hyperref}

\title{Multi-label feature selection based on binary hashing learning and dynamic graph constraints

}

\author{
  Cong Guo \\
  Key Laboratory of Intelligent Education Technology and Application of Zhejiang Province\\Zhejiang Normal University,Jinhua, China \\
  \texttt{guocong@henu.edu.cn;} \\
  \AND
  Changqin Huang*\\
  Key Laboratory of Intelligent Education Technology and Application of Zhejiang Province\\Zhejiang Normal University,Jinhua, China \\
  \texttt{cqhuang@zju.edu.cn;} \\
  \AND
    Wenhua Zhou\\
    Key Laboratory of Intelligent Education Technology and Application of Zhejiang Province\\Zhejiang Normal University,Jinhua, China \\
\AND
Xiaodi Huang\\
  School of Computing, Mathematics and Engineering, Charles Sturt University,Albury,Australia \\
   \\
}

\begin{document}
\maketitle

\begin{abstract}
Multi-label learning poses significant challenges in extracting reliable supervisory signals from the label space. Existing approaches often employ continuous pseudo-labels to replace binary labels, improving supervisory information representation. However, these methods can introduce noise from irrelevant labels and lead to unreliable graph structures. To overcome these limitations, this study introduces a novel multi-label feature selection method called Binary Hashing and Dynamic Graph Constraint (BHDG), the first method to integrate binary hashing into multi-label learning. BHDG utilizes low-dimensional binary hashing codes as pseudo-labels to reduce noise and improve representation robustness. A dynamically constrained sample projection space is constructed based on the graph structure of these binary pseudo-labels, enhancing the reliability of the dynamic graph. To further enhance pseudo-label quality, BHDG incorporates label graph constraints and inner product minimization within the sample space. Additionally, an $l_{2,1}$-norm regularization term is added to the objective function to facilitate the feature selection process. The augmented Lagrangian multiplier (ALM) method is employed to optimize binary variables effectively. Comprehensive experiments on 10 benchmark datasets demonstrate that BHDG outperforms ten state-of-the-art methods across six evaluation metrics. BHDG achieves the highest overall performance ranking, surpassing the next-best method by an average of at least 2.7 ranks per metric, underscoring its effectiveness and robustness in multi-label feature selection.
\end{abstract}


\section{Introduction}
In modern scientific domains such as genomics, text analysis, and image processing, data often exhibit high-dimensional characteristics \cite{A1}. While high-dimensional data provide valuable information, they also pose significant challenges, especially the “curse of dimensionality” \cite{A2}. High dimensionality increases model complexity and slows the training process of algorithms. To address these issues, feature selection has been widely applied as a critical dimensionality reduction method \cite{A3}, aiming to identify features most relevant to the prediction task while minimizing information loss \cite{A4}.

Existing feature selection algorithms are generally categorized into three types based on their relationship with the classifier: wrapper, filter, and embedded methods \cite{A5,A6}. Filter methods are independent of machine learning classifiers, wrapper methods rely on classifier performance, and embedded methods integrate feature selection directly into the classifier's learning process, treating feature selection as an optimization problem. Among these, embedded algorithms usually exhibit better performance with lower computational complexity, making them a focus of recent research. Therefore, this paper focuses on advancing embedded feature selection algorithms.

In the field of multi-label feature selection, extracting useful information and correlations from the original labels to enhance feature selection remains a critical challenge. Some studies have proposed learning numerical pseudo-labels to replace the original binary labels, enabling better utilization of supervisory information. For example, Zhang et al.\cite{A7} developed a non-negative multi-label feature selection method (NMDG) that uses the dynamic graph structure of pseudo-labels to constrain feature weights and explore label correlations. Similarly, Zhang et al.\cite{A8} introduced a multi-label feature selection method based on latent labels and dynamic graph constraints (LRDG), which learns numerical pseudo-labels while using their graph structure to constrain feature weights. Despite their success, these pseudo-label-based methods face two major issues: 

{\bf{Noise Information}}: Figure 1 effectively highlights this issue. The original labels associated with the image include "human," "shoe," "car," "river," and "grass." Among these, "human," "river," and "grass" are the most prominent and significant. Conversely, the label "cat" is irrelevant since no animals are present in the image. However, existing methods often generate pseudo-labels that mistakenly include "cat" due to its presence in the original label space, thereby reducing the model's generalizability.

In contrast, the binary pseudo-labels shown in Figure 1 offer a more refined label representation. For instance, "cat" is categorized under an irrelevant class, "animals," while "river," "car," and "grass" are grouped under "scenery." These low-dimensional binary pseudo-labels effectively reduce noise through several mechanisms.

First, binary labels eliminate ambiguity by assigning clear-cut values (0 or 1) to classes, avoiding the uncertainty introduced by the intermediate probabilities of continuous pseudo-labels. This prevents the model from misinterpreting low-confidence predictions as useful information. Second, binary labels inherently simplify the label space by grouping similar or irrelevant labels into broader, more meaningful categories, such as consolidating "river," "car," and "grass" under "scenery." This simplification reduces the risk of overfitting to minor, inconsequential variations. Third, binary pseudo-labels enforce a sharper distinction between relevant and irrelevant labels, enabling the model to concentrate on learning salient features instead of being distracted by noisy or redundant information.

By assigning a value of 0 to irrelevant labels like "animals," binary pseudo-labels ensure compactness and clarity, allowing our approach to better distinguish between relevant categories. Thus, using such binary latent labels as supervisory signals enhances the model's robustness by effectively minimizing noise, filtering irrelevant information, and fostering efficient learning of essential label structure.
\begin{figure}
    \centering
    \includegraphics[width=1\linewidth]{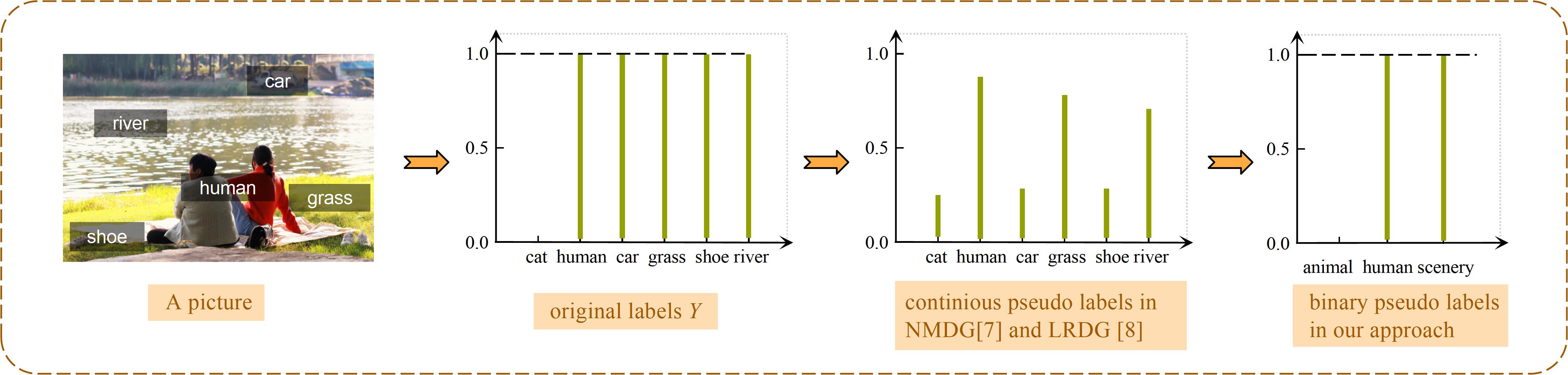}
    \caption{An example of continuous and binary pseudo-label learning.}
    \label{fig:enter-label}
\end{figure}

{\bf{Unreliable graph structure}}: Some algorithms such as NMDG\cite{A7} and LRDG\cite{A8} use the graph structure of pseudo labels to constrain other variables to guide feature selection. However, since these pseudo labels introduce noise information, this graph structure cannot reflect the actual data distribution, resulting in the accumulation of training errors. Therefore, in view of the above analysis, it is necessary to learn a binary low-dimensional embedding and refine pseudo-label learning based on its graph structure.

To address these limitations, we introduce binary hashing learning, a method widely applied in fields such as information retrieval \cite{A10}. Hashing learning projects high-dimensional data into compact binary codes (0 and 1), preserving both similarity structure and semantic information \cite{A11}. By leveraging the strengths of hashing learning, the generated binary hashing codes can serve as latent labels in pseudo-label learning, effectively filtering out noise while retaining the core information required for the classification task.

Building on these insights, we introduce a novel multi-label feature selection method called BHDG. This method projects original labels into low-dimensional binary pseudo-labels to reduce noise and preserve essential information. The graph structure of these pseudo-labels is then used to dynamically constrain the sample projection space, ensuring robust pseudo-label learning. BHDG further incorporates original label graph constraints and inner product minimization within the sample space to capture high-quality binary pseudo-labels. These mechanisms enable the model to align effectively with the binary pseudo-labels, mitigating noise and minimizing training errors through dynamic constraints. The refined binary pseudo-labels are then utilized to guide the feature selection process. All these components are seamlessly integrated into a unified objective function. To optimize this function, we develop an efficient discrete optimization approach. This comprehensive framework ensures that BHDG achieves both accuracy and robustness in multi-label feature selection. The main contributions of this paper are as follows.

\begin{itemize}
    \item {\bf{Binary Pseudo-Label Learning}}: Existing pseudo-label methods often neglect binary forms. To address this, we introduce a novel multi-label feature selection method, BHDG, which incorporates binary latent labels to enhance pseudo-label learning.

    \item {\bf{Preservation of Semantic Information}}: The proposed algorithm retains original semantic information by employing inner product minimization of the similarity matrix and label graph constraints. It also dynamically constrains the sample projection space using a hashing label graph structure, facilitating smooth information transfer between the labels and sample spaces while preventing critical supervisory signals.

    \item {\bf{Efficient Optimization}}: We use a discrete optimization approach to solve the objective function and provide theoretical proof of its convergence. Experimental results confirm the method's efficiency and reliable convergence. 

    \item {\bf{Comprehensive Performance Evaluation}}: BHDG was rigorously evaluated on 10 multi-label datasets across six performance metrics. Experimental results demonstrate that BHDG outperforms 10 state-of-the-art methods, validating its effectiveness and superiority in multi-label feature selection.

\end{itemize}

\begin{table}
\renewcommand\arraystretch{1.3}
\caption{The description of notations.}
    \resizebox{\linewidth}{!}{
    \begin{tabular}{ll}
    \hline
       Symbol  &Definition \\
    \hline
        $A_{.i}$ &The $j$-th column of $A$. \\
        $A_{i.}$ &The $i$-th row of $A$. \\
        $A_{ij}$ &The ($i,j$)-th element of $A$. \\
        $A^{T}$ &The transpose of $A$. \\
        $A^{-1}$ &The inverse of $A$. \\
       $tr(A)$	 &The trace of $A$. \\
        ${\left\| A \right\|_F}$ &For matrix $A \in {\Re ^{n \times d}}$, ${\left\| A \right\|_F} = \sqrt {\sum\nolimits_{i = 1}^n {\sum\nolimits_{j = 1}^d {A_{ij}^2} } }$denotes the Frobenius norm of $A$. \\
        ${\left\| A \right\|_{2,1}}$ & For matrix $A \in {\Re ^{n \times d}}$,${\left\| A \right\|_{2,1}} = \sum\nolimits_{i = 1}^n {\sqrt {\sum\nolimits_{j = 1}^d {A_{ij}^2} } } $denotes the $l_{2,1}$ norm of $A$.\\
    \hline
    \end{tabular}}
    
    \label{tab:my_label}
\end{table}

\section{Related work}
In this subsection, we review the relevant literature. Prior to this, we first introduce the symbols and definitions used in this paper, which are summarized in Table 1. The objective of the feature selection task is to learn a weight matrix (also called a projection matrix) $W \in R^{d \times c}$, where the $l_{2}$-norm of the $i$-th row of $W$ represents the importance of the $i$-th feature.
In multi-label feature selection, a common approach is to use sparse regression as the fundational framework, and then  embed a graph regularization term to constrain the manifold structure of the variables\cite{A12}. Given a data matrix $X \in R^{n×d}$ and the corresponding label matrix $Y \in \{0,1\}^{n×c}$, the basic formulation of this task for learning the projection matrix $W$, is expressed as follows:
\begin{equation}
    \mathop {\min }\limits_W \left\| {XW - Y} \right\|_F^2 + \alpha R(W) + \lambda G(W),
\end{equation}
where, $R(W)$ is the regularization term that helps mitgate overfitting, and $G(\cdot)$ represents the graph constraint term. 

Graph-based constraints are widely used in multi-label feature selection tasks to capture the relationships between features or labels. These constraints can generally be classified into three categories: fixed graph constraints, dynamic graph constraints, and adaptive graph constraints. Each type of graph constraint has its own strengths and focuses. Below, we will briefly review the literature on each of these approaches.

\subsection{Fixed graph constraints}
Fixed graph constraints are the most commonly used approach in manifold learning. These constraints construct a fixed adjacency graph based on the original data before learning the manifold space. The edge weights in the graph are determined by the similarity measure between two nodes \cite{A13}. Importantly, the adjacency graph remains unchanged throughout the feature selection process. 

For instance, in the problem described in Eq.(1), the graph constraint can be formulated as follows:
\begin{equation}
    \mathop {\min }\limits_W \sum\limits_{i = 1}^c {\sum\limits_{j = 1}^c {{S_{ij}}({{\left( {XW} \right)}_{i.}} - {{\left( {XW} \right)}_{j.}})} }  = Tr({W^T}{X^T}{L_Y}XW),{\rm{ }}
\end{equation}

where the graph Laplacian matrix ${L_Y}={A_Y}-{S_Y}$ is the graph Laplacian matrix calculated on the label matrix $Y$, and ${\left( {{S_Y}} \right)_{ij}} = \exp \left( { - \frac{{\left\| {{Y_{.j}} - {Y_{.i}}} \right\|_2^2}}{\sigma }} \right),$if and only if $Y_{.i}$ and $Y_{.j}$ are neighbors. Here, $A_{Y}$ is a diagonal matrix, whose diagonal elements ${\left( {{A_Y}} \right)_{ii}} = \sum\limits_{j = 1}^c {{{\left( {{S_Y}} \right)}_{ij}}}$. The geometric interpretation of Eq.(2) is that the graph structure of the label matrix $Y$ is used to constrain the dimensionality reduction projection space $XW$.  

Feature selection based on fixed graph constraints has achieved significant success. For example, Jian \cite{A14} used instance graph constraints on the manifold structure of latent semantics, leading to the development of the MIFS algorithm; Li \cite{A15} proposed MSFS which incorporates instance graph constraints on the structure of the dimensionality reduction space and a joint sparse regularization term to enhance the robustness of the objective function; Li \cite{A16} introduced an algorithm based on self-expressive labels, which uses the original label graph to constrain the structure of self-expressive labels, enabling the exploration of label correlations.

While fixed graph constraints offer relatively low computational overhead and good stability, their primary limitation is that the graph structure remains static throughout the process. As a result, the performance of the algorithm is heavily dependent on the quality of the graph, and it struggles to effectively capture the global structure of the data. 

\subsection{Dynamic graph constraints}
Dynamic graph constraints are similar to fixed graph constraints in terms of graph construction but differ in that the graph is recalculated during the iterative process. Taking Eq.(1) as an example, we assume that the projection matrix $W$ shares a similar structure with the dimensionality-reduced space $XW$:
\begin{equation}
    \mathop {\min }\limits_W \sum\limits_{i = 1}^c {\sum\limits_{j = 1}^c {{S_{ij}}({{\left( {XW} \right)}_{.i}} - {{\left( {XW} \right)}_{.j}})} }  = Tr({W^T}{X^T}{L_W}XW),{\rm{  }}
\end{equation}
Where the calculation of LW is similar to that of $L_Y$ in Eq.(2). However, unlike fixed graph constraints, $L_W$ needs to be recalculated in each iteration based on the most recent $W$ value. This dynamic recalculation allows for a more flexible representation of the manifold structure as the algorithm progresses.

Feature selection based on dynamic graph constraints has gained considerable attention in recent years. For example, Zhang \cite{A7} proposed NMDG, which uses pseudo-labels to replace original labels and applies dynamic graph constraints to the projection matrix of the pseudo-labels to mitigate the information loss typically observed in static graph methods.  Hu \cite{A9} proposed DSMFS, which uses graph regularization to explore label correlations while utilizing dynamic latent representation graph constraints to improve feature selection performance. Zhang \cite{A8} proposed LRDG, which also employs numerical pseudo-labels and dynamically constrained pseudo-label graphs to replace original labels, improving the learning of supervisory information through the dynamic graph structure. 

Dynamic graph constraints are beneficial in adapting to complex manifold structures, as they allow the graph to evolve with each update of the variables. This results in better robustness compared to fixed graph constraints. However, the need to recalculate the graph at each iteration increases computational overhead, making this approach computationally more expensive.

\subsection{Adaptive graph constraints}
Adaptive graph constraints differ from both fixed and dynamic graph constraints in that they embed the graph matrix directly into the model framework, eliminating the need for explicit computation of the nearest-neighbor relationships\cite{A17}. For example, the problem in Eq.(2) can be transformed into an adaptive graph constraint, represented as follows:
\begin{equation}
    \begin{array}{l}
\mathop {\min }\limits_{W,S} \sum\limits_{i,j} {\left( {{s_{ij}}\left\| {{Y_{i.}} - {Y_{j.}}} \right\|_2^2 + \alpha s_{ij}^2} \right)}  + tr\left( {{W^T}{X^T}{L_S}XW} \right),{\rm{ }}\\
{\rm{s}}{\rm{.t}}{\rm{. }}0 \le {s_{ij}} \le 1,\forall i,s_i^T{\bf{1}} = 1.{\rm{ }}
\end{array}
\end{equation}
where $L_{S}=D-(S+S^{T})/2$ is the graph Laplacian matrix, $D$ is a diagonal matrix, whose diagonal elements ${D_{ii}} = \sum\nolimits_j {\left( {\left( {{s_{ij}} + {s_{ji}}} \right)/2} \right)}$, $\alpha$ is the regularization parameter. In Eq.(4), the similarity matrix $S$ is treated as a variable that needs to be optimized.

In recent years, multi-label learning methods based on adaptive graphs have attracted attention. For example, Qin \cite{A18} proposed  AGLE, which combines adaptive graph learning with label information enhancement. This method learns the label similarity matrix adaptively and uses it to constrain the projection space, enhancing label information and leading to more robust results. Ma \cite{A19} proposed MFS-AGD, which employs adaptive graph diffusion. This method learns an adaptive graph from pseudo-labels and uses this graph to constrain the structure of the feature space, thereby exploring the higher-order structure and supervisory information more effectively.

Adaptive graph constraints offer better flexibility compared to fixed and dynamic graphs \cite{A20}, as they can better balance local and global structures within the data. However, learning the adaptive graph structure is relatively complex, and no universally established optimization approach exists. Furthermore, since the graph structure is treated as a variable that needs to be optimized, exploring multiple adaptive graph structures can significantly increase computational complexity.
\section{The proposed BHDG method}
In order to perform binary pseudo-label learning, we first use sparse regression as the basic framework and then project the original labels into low-dimensional binary pseudo-labels as supervision information. Based on dynamic graph constraints, label graph constraints, and the minimization of the inner product, we simultaneously guide the feature selection while constraining the structure of the hashing matrix. The overall framework of the algorithm is shown in Figure 2, and the specific steps are as follows:

\begin{figure}
    \centering
    \includegraphics[width=1\linewidth]{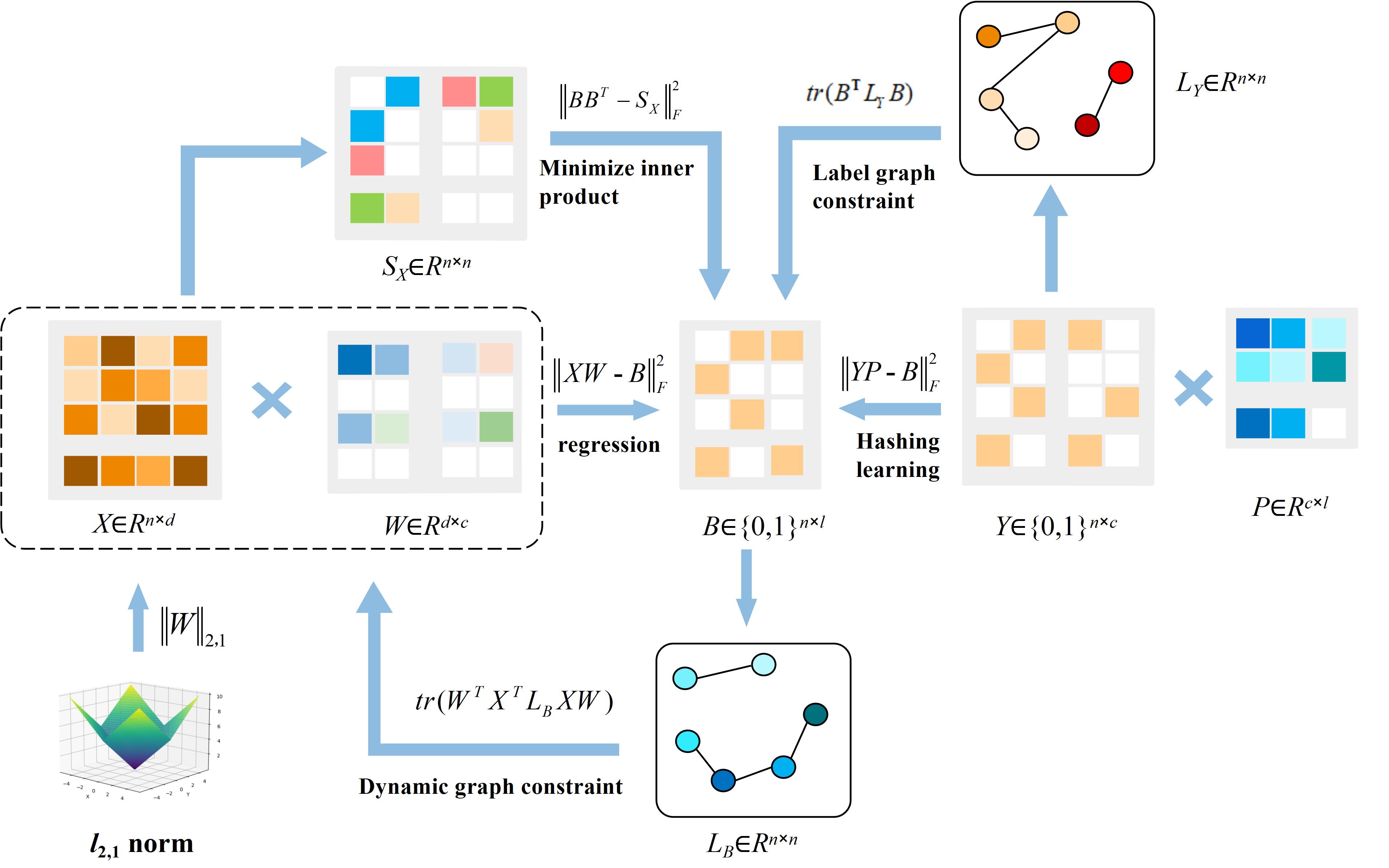}
    \caption{The overall framework of the proposed BHDG. In this framework, $B$ represents the binary hashing labels.}
    \label{fig:enter-label}
\end{figure}

\subsection{Basic model}
The proposed algorithm integrates the least squares loss function, sparse term, graph constraint term, and hashing learning term to form a unified objective function. The loss function captures the relationship between the data and labels. The sparse term ensures that the variables remain sparse, reducing overfitting risks. The graph constraint term learns the manifold structure of the data, capturing complex latent semantics. The hashing learning term generates the binary hash matrix and provides supervisory signals for efficient feature selection. The basic model can be formulated as:

\begin{equation}
    \begin{array}{l}
\Theta  = \mathop {\min }\limits_{W,B,P} Q(B,W) + R(W) + M(W,P,B) + H(W,P,B),\\
s.t.{\rm{ }}\left\{ {W,B,P} \right\} \ge 0,B \in {\{ 0,1\} ^{n \times l}}.
\end{array}
\end{equation}

In Eq.(5), the term $Q(\cdot)$ represents the least squares loss function; the second term $R(\cdot)$ represents the regularization term; the third term $M(\cdot)$ represents the graph constraint term for the hash matrix; and the fourth term $H(\cdot)$ represents the hashing learning term.

For $Q(\cdot)$, it is used to fit the labels. By projecting the original label space into the binary low-dimensional embeddings, the influence of noise is suppressed,  and feature selection is guided more effectively. We formulize $Q(\cdot)$ as:
\begin{equation}
    \begin{array}{l}
Q(W,B) = \left\| {XW - B} \right\|_F^2, s.t.{\rm{ }}B \in {\{ 0,1\} ^{n \times l}}.
\end{array}
\end{equation}
Here, $X \in R^{n\times d}$ is an input data matrix with $n$ samples and $d$ features, $W \in R^{d\times l}$ is the feature weight matrix, quantifying feature importance. $B \in R^{n\times l}$ is a binary hashing matrix that preserves the semantic structure in the dimensionality-reduced space. For $R(\cdot)$, we use the $l_{2,1}$-norm to encourage sparsity in $W$ :
\begin{equation}
    R(W) = {\lambda _1}{\left\| W \right\|_{2,1}} = {\lambda _1}\sum\nolimits_{i = 1}^n {\sqrt {\sum\nolimits_{j = 1}^l {W_{ij}^2} } } .
\end{equation}
Where $\lambda_1$ is the regularization parameter. By introducing the $l_{2,1}$-norm constraint on the weight matrix $W$ in the least squares regression model, it encourages row sparsity by selecting a small number of key features. The $l_{2,1}$-norm also enhances the robustness of the model, reducing sensitivity to noise and outliers.

By integrating these terms, the BHDG method effectively aligns feature selection with binary hashing label learning, achieving both noise reduction and meaningful feature selection.
\subsection{Binary hashing learning}
Continuous numerical pseudo-labels often introduce noise from irrelevant labels, making them less effective for tasks requiring robust label representations. To mitigate this issue, we propose learning binary pseudo-labels, which inherently reduce noise while preserving critical semantic information. The binary pseudo-label learning process is defined as: 
\begin{equation}
    \begin{array}{l}
\mathop {\min }\limits_B \left\| {YP - B} \right\|_F^2,
s.t.B \in {\left\{ {0,1} \right\}^{n \times l}}.
\end{array}
\end{equation}
Where $P \in R^{c\times l}$ is a projection matrix used to align the relationship between the original label matrix $Y$ and the binary label matrix $B$. 

Inspired by Wang et al. \cite{A21}, binary labels are designed to preserve the semantic similarity of instances. Specifically, semantically similar data points should have similar binary labels. This principle can be formulated as the following inner product minimization problem:
\begin{equation}
    \mathop {\min }\limits_B \left\| {B{B^T} - {S_X}} \right\|_F^2.
\end{equation}
Where $S_X$ is the similarity matrix for $X$, computed using a kernel function:
\begin{equation}
    {({S_X})_{ij}} = \left\{ \begin{array}{l}
{e^{ - \frac{{\left\| {{X_{i.}} - {X_{j.}}} \right\|_2^2}}{\sigma }}},if{\rm{ }}{X_{i.}} \in {N_k}\left( {{X_{j.}}} \right){\rm{ }}or{\rm{ }}{X_{j.}} \in {N_k}\left( {{X_{i.}}} \right).\\
0,{\rm{           }}others.
\end{array} \right.
\end{equation}
Here, $X_{i.}$ and $X_{j.}$ are two samples, and $\sigma$ controls the scale of the Gaussian kernel.
By combining Eqs.(8) and (9), the hashing learning term $H(\cdot)$ can be expressed as:
\begin{equation}
    \begin{array}{l}
H\left( {P,B} \right) = \left\| {YP - B} \right\|_F^2 + \left\| {B{B^T} - {S_X}} \right\|_F^2,\\
s.t.B \in {\left\{ {0,1} \right\}^{n \times l}}.
\end{array}
\end{equation}

\subsection{Label graph constraint}
Since $B$ represents a binary latent representation label matrix, it should preserve the semantic information of the original supervisory information $Y$. To ensure this consistency, we assume that $B$ and the original label $Y$ share a similar geometric structure. Formally, if $Y_{i.}$ and $Y_{j.}$ are similar, then $B_{i.}$ and $B_{j.}$ should also exhibit similarity. To formalize this idea, we introduce the following constraint:

\begin{equation}
    \frac{1}{2}\sum\limits_{i = i}^n {\sum\limits_{j = 1}^n {{{({S_Y})}_{ij}}\left\| {{B_{i.}} - {B_{j.}}} \right\|_2^2} }  = tr({B^{\bf{T}}}{L_Y}B).
\end{equation}
where $S_Y$ is the similarity matrix for $Y$, capturing the semantic relationships among the labels.

This constraint ensures that the binary latent representation B retains the geometric structure of $Y$, aligning semantic information across the two spaces.
\subsection{Dynamic graph constraint}
Graph regularization has been widely adopted to enforce structural consistency between different spaces, allowing them to share more meaningful information. Building on this idea, we incorporate dynamic graph constraints to further improve structural alignment between the instance projection space and the binary hashing space. For the assumption in Eq.(9), we construct a dynamic graph Laplacian matrix for the hashing matrix $B$. Formally, this is expressed as:
\begin{equation}
    \frac{1}{2}\sum\limits_{i,j = 1}^n {\left\| {{x_i}W - {x_j}W} \right\|_2^2} {\left( {{S_B}} \right)_{ij}} = tr({W^T}{X^T}{L_B}XW).
\end{equation}
Here, $L_{B} = A_{B}-S_{B}$ is the graph Laplacian matrix for $B$, and ${\left( {{A_B}} \right)_{ii}} = \sum\limits_{j = 1}^n {{{\left( {{S_B}} \right)}_{ij}}} $, $S_B$ is the similarity matrix for $B$, computed using the cosine similarity:
\begin{equation}
    {({S_B})_{ij}} = \left\{ \begin{array}{l}
\frac{{B_{i.}^T{B_{j.}}}}{{{{\left\| {{B_{i.}}} \right\|}_2}{{\left\| {{B_{j.}}} \right\|}_2}}},if{\rm{ }}{B_{i.}} \in {N_k}\left( {{B_{j.}}} \right){\rm{ }}or{\rm{ }}{B_{j.}} \in {N_k}\left( {{B_{i.}}} \right).\\
0,{\rm{           }}others.
\end{array} \right.
\end{equation}
Here, $B_{i.}$ and $B_{j.}$ are the binary representations of instances $i$ and $j$, respectively.

In Eq.(13), The dynamic adjustment of graph Laplacian $L_B$ ensures that the graph Laplacian becomes increasingly reliable over time, reducing error propagation. On this basis, the graph constraint term $M(\cdot)$ is expressed as follows:

\begin{equation}
    M\left( {W,B} \right) = {\lambda _2}tr({W^T}{X^T}{L_B}XW) + {\lambda _3}tr({B^{\bf{T}}}{L_Y}B).
\end{equation}
where $\lambda_2$ and $\lambda_3$ are positive balance factors that control the influence of different manifold regularization terms. This formulation enables smooth and robust information transfer across dimensionality-reduced space, label space, and binary hashing space. By integrating these two graph constraints, $M(\cdot)$ enhances the structural alignment, ensuring effective feature selection and accurate pseudo-label learning.
\subsection{Overall objective function}
Building on the discussions in the previous sections, the overall objective function of the proposed algorithm is summarized as: 
\begin{equation}
    \begin{array}{l}
\Theta (W,P,B) = \underbrace {\left\| {XW - B} \right\|_F^2}_{regression} + \underbrace {\left\| {YP - B} \right\|_F^2}_{\text{hashing{\rm{ }}learning}} + \underbrace {{\lambda _1}{{\left\| W \right\|}_{2,1}}}_{{\rm{regularization}}} + \underbrace {{\lambda _2}tr({W^T}{X^T}{L_B}XW)}_{\text{dynamic{\rm{ }}graph{\rm{ }}constraint} } + \\
\underbrace {{\lambda _3}tr({B^T}{L_Y}B)}_{label\;graph\;{\rm{ }}constraint } + \underbrace {\left\| {B{B^T} - {S_X}} \right\|_F^2}_{\text{minimize{\rm{ }}inner{\rm{ }}product}},\\
s.t.\left\{ {W,P,B} \right\} \ge 0,B \in {\left\{ {0,1} \right\}^{n \times l}}.
\end{array}
\end{equation}

The proposed algorithm integrates binary pseudo-label learning, semantic preservation, and graph regularization into a unified framework. By jointly optimizing these terms, the model effectively learns latent binary hashing labels while preserving the semantic and structural consistency of the original supervisory information. This comprehensive approach allows for a more comprehensive and accurate representation of the data, reducing information loss and improving the quality of feature weights. As a result, the algorithm achieves superior performance in multi-label feature selection, as validated in the experiments.

\section{Solution}
In this section, we present the optimization scheme of the proposed algorithm in Section 4.1, followed by showing the computational complexity of the algorithm in Section 4.2, and finally, discussing its convergence in Section 4.3.
\subsection{Optimization schemes}
In this section, we propose a simple yet effective optimization method to solve the non-convex objective function in Eq.(16). To tackle the inherent non-convexity, we decompose the objective function into multiple sub-problems and solve them sequentially by updating one variable while keeping the others fixed. The objective function (16) is equivalent to:
\begin{equation}
    \begin{array}{l}
\mathop {\min }\limits_{W,P,B} tr\left( {{{\left( {XW - B} \right)}^T}\left( {XW - B} \right)} \right) + tr\left( {{{\left( {YP - B} \right)}^T}\left( {YP - B} \right)} \right) + {\lambda _1}tr\left( {{W^T}DW} \right) + \\
{\lambda _2}tr({W^T}{X^T}{L_B}XW) + {\lambda _3}tr({B^T}{L_Y}B) + tr\left( {{{\left( {B{B^T} - {S_X}} \right)}^T}\left( {B{B^T} - {S_X}} \right)} \right).\\
s.t.\left\{ {W,P,B} \right\} \ge 0,B \in {\left\{ {0,1} \right\}^{n \times l}}.
\end{array}
\end{equation}
Where D is a diagonal matrix whose diagonal elements are expressed as: ${D_{ii}} = \frac{1}{{2{{\left\| {{w_{i.}}} \right\|}_2} + \varepsilon }}$,$\varepsilon$ is a small non-negative constant.
\subsubsection{Optimizing $W$ and $P$}
To optimize $W$ and $P$ , we first construct Lagrangian function with respect to $W$ and $P$ as follows:
\begin{equation}
    \begin{array}{l}
\Theta (W,P) = tr\left( {{{\left( {XW - B} \right)}^T}\left( {XW - B} \right)} \right) + tr\left( {{{\left( {YP - B} \right)}^T}\left( {YP - B} \right)} \right) + \\
{\lambda _1}tr\left( {{W^T}DW} \right) + {\lambda _2}tr({W^T}{X^T}{L_B}XW) - tr(\Psi {W^T}) - tr(\Phi {P^T}),\\
s.t.\left\{ {W,P} \right\} \ge 0.
\end{array}
\end{equation}
Where $\Psi  \in {\Re ^{d \times l}}$ and $\Phi  \in {\Re ^{c \times l}}$ are Lagrange multipliers. By taking partial derivatives of $W$ and $P$ using Eq.(18) and setting them to 0, we can obtain: 
\begin{equation}
    \left\{ \begin{array}{l}
\frac{{\partial \Theta }}{{\partial W}} = 2{X^T}XW - 2{X^T}B + 2{\lambda _2}{X^T}{L_B}XW + 2{\lambda _1}DW - \Psi .\\
\frac{{\partial \Theta }}{{\partial P}} = 2{Y^T}YP - 2{Y^T}B - \Phi .
\end{array} \right.
\end{equation}
Through the complementary slackness of the Karush-Kuhn-Tucker (KKT) condition, we have ${\Psi _{ij}}{W_{ij}} = 0$ and ${\Phi _{ij}}{P_{ij}} = 0$. We can get:
\begin{equation}
    \left\{ \begin{array}{l}
\left( {{X^T}XW - {X^T}B + {\lambda _2}{X^T}{L_B}XW + {\lambda _1}DW} \right) \circ W = 0.\\
\left( {{Y^T}YP - {Y^T}B} \right) \circ P = 0.
\end{array} \right.
\end{equation}
Where $\circ$ represents the Hadamard product of two matrices. In Eq.(20), we transform the matrix $L_B$ into the form of $A_B-S_B$, and further obtain the update formulas of $W$ and $P$:
\begin{equation}
    \left\{ \begin{array}{l}
W_{ij}^{^{t + 1}} \leftarrow W_{ij}^t\frac{{{{\left( {{X^T}B + {\lambda _2}{X^T}{S_B}X{W^t}} \right)}_{ij}}}}{{{{\left( {{X^T}X{W^t} + {\lambda _2}{X^T}{A_B}X{W^t} + {\lambda _1}D{W^t}} \right)}_{ij}} + \varepsilon }}.\\
P_{ij}^{^{t + 1}} \leftarrow P_{ij}^t\frac{{{{\left( {{Y^T}B} \right)}_{ij}}}}{{{{\left( {{Y^T}Y{P^t}} \right)}_{ij}} + \varepsilon }}.
\end{array} \right.
\end{equation}
Where $t$ is the iteration times and $\varepsilon $ is a small non-negative constant that can prevent perturbation of non-differentiable and the zero-denominator problems.

\subsubsection{Optimizing $B$}
When fixing the other variables, the optimization problem for $B$ becomes:
\begin{equation}
    \begin{array}{l}
\mathop {\min }\limits_B \left\| {B{B^T} - {S_X}} \right\|_F^2 - 2tr({B^T}XW) + 2tr({B^T}B) - 2{\lambda _2}tr({B^T}YP) + {\lambda _3}tr({B^T}{L_Y}B)\\
s.t.B \in {\left\{ {0,1} \right\}^{n \times l}}.
\end{array}
\end{equation}

Due to the discrete constraints on the binary label B, directly solving Eq.(22) is an NP-hard problem. Inspired by Shi \cite{A22}, we employ a discrete optimization method based on the augmented Lagrangian multiplier (ALM) to solve for B. Specifically, for Eq.(22), we introduce an auxiliary discrete variable $Z\in{0, 1}_{n×l}$ to replace the rightest B in each term, while maintaining their equivalence during the optimization process. This leads to the following optimization formula:
\begin{equation}
    \begin{array}{l}
\mathop {\min }\limits_{B,Z \in {{\left\{ {0,1} \right\}}^{n \times l}},M} \left\| {B{Z^T} - {S_X}} \right\|_F^2 - 2tr({B^T}XW) - 2{\lambda _2}tr({B^T}YP) + 2tr({B^T}Z) + \\
{\lambda _3}tr({B^T}{L_Y}Z) + \frac{\rho }{2}\left\| {B - Z + \frac{M}{\rho }} \right\|_F^2.
\end{array}
\end{equation}
where $M\in R^{n \times l}$ represents the Lagrange multiplier. 

{\bf{Updating $B$}}: Taking the partial derivative of Eq.(23) with respect to $B$ and setting it to zero yields the closed-form solution for $B$:
\begin{equation}
    B = \left( {{\mathop{\rm sgn}} \left( {\left( {2XW + 2{S_X}Z + 2{\lambda _2}YP - {\lambda _3}{L_Y}Z - M + (\rho  - 2)Z} \right){{\left( {2{Z^T}Z + \rho {I_l}} \right)}^{ - 1}}} \right) + 1} \right)/2
\end{equation}
where $sgn$ is the sign function, and $I_l$ is the identity matrix of size $l \times l$.

{\bf{Updating $Z$}}:When fixing other variables, we take the partial derivative of Eq.(23) with respect to $Z$ and set it to zero, yielding the analytical solution for $Z$:
\begin{equation}
    Z = \left( {{\mathop{\rm sgn}} \left( {\left( {2{{\left( {{S_X}} \right)}^T}B - {\lambda _3}{L_Y}B + (\rho  + 2)B + M} \right){{(2{B^T}B + \rho {I_l})}^{ - 1}}} \right) + 1} \right)/2.
\end{equation}
{\bf{Updating $M$ and $\rho$}}: According to the ALM theorem, the update formulas for $M$ and $\rho$ are given by:
\begin{equation}
    M = M + \rho \left( {B - Z} \right),\rho  = \alpha \rho .
\end{equation}
where $\alpha$ is a balancing factor used to adjust the step size. The pseudocode for the BHDG algorithm is summarized in Algorithm 1. Where the variables $W$, $P$, $B$, $Z$, and $M$ are updated iteratively until convergence.

\label{sec1}

\begin{algorithm}[h]
\caption{BHDG for Feature Selection}\label{alg:two}
\KwIn{Data matrix $X \in {\Re ^{n \times d}}$, label matrix $Y \in {\Re ^{n \times c}}$, parameters ${\lambda}_{1},{\lambda}_{2},{\lambda}_{3},\rho,\alpha$.}
\KwOut{Feature weighting vector.}
$t=0$;\\
Initialize $W^t$,$P^t$, and $M$ as random matrices;\\
Initialize $B^t$ and $Z$ as two binary random matrices;\\
Calculate $L_B$ on $B^t$, and $L_Y$ on $Y$, respectively;\\
\Repeat{The objective function is converged}{
$t=t+1$;\\
Update ${D_{ii}}$ by $\frac{1}{{2{{\left\| {{w_{i.}}} \right\|}_2} + \varepsilon }}$.\\
Compute $W^t$ and $P^t$ according to Eq.(21).\\
Compute $B^t$ according to Eq.(24).\\
Update $Z$ according to Eq.(25).\\
Update $M$ and $\rho$ according to Eq.(26).\\
Recalculate $L_B$ on $B^t$;\\
}
Calculate ${\bf{v}}={\left\| {\left( {W^t} \right)_{i.}} \right\|_2}(i = 1,2,...,d)$;\\

{\bf{Return}} feature importance vector ${\bf{v}}$.
\end{algorithm}

\subsection{Computational complexity analysis}
For Algorithm 1, in each iteration, the partial derivatives of the matrices \( W \), \( P \), \( B \), and \( Z \) need to be computed once, with their respective computational complexities being \( O(ndl + dn^2 + d^2n + d^2c) \), \( O(c^2n + ncl + c^2l) \), \( O(n^2l + ndc + l^3 + ncl) \), and \( O(n^2l + l^3) \). Considering that the computational cost of updating \( M \) is negligible, the overall computational complexity of the algorithm is \( O(t(ndl + dn^2 + d^2n + d^2c + c^2n + ncl + c^2l + n^2l + ndc + l^3)) \), where \( t \) denotes the number of iterations of the algorithm.

\subsection{Proof of convergence}
In this subsection, we provide a proof of the convergence of the proposed optimization method. We first present the proof for the variables $P$ and $W$. Taking the variable $P$ as an example, we derive its following update rule by using the gradient descent method:
\begin{equation}
    P_{ij}^{t + 1} \leftarrow P_{ij}^t - \eta {\left( {\frac{{\partial \Theta }}{{\partial {P^t}}}} \right)_{ij}}
\end{equation}
where the learning rate $\eta$ is a positive constant. To adaptively adjust the learning rate, we set:
\begin{equation}
    \eta  = \frac{{P_{ij}^t}}{{2{{\left( {{Y^T}Y{P^t}} \right)}_{ij}}}}.
\end{equation}
Substituting Eq.(28) into Eq.(27), we obtain the following result:
\begin{equation}
    P_{ij}^{t + 1} \leftarrow P_{ij}^t - \frac{{P_{ij}^t}}{{2{{\left( {{Y^T}Y{P^t}} \right)}_{ij}}}}{\left( {\frac{{\partial \Theta }}{{\partial {P^t}}}} \right)_{ij}} \Leftrightarrow P_{ij}^{t + 1} \leftarrow P_{ij}^t\frac{{{{\left( {{Y^T}B} \right)}_{ij}}}}{{{{\left( {{Y^T}Y{P^t}} \right)}_{ij}}}}.
\end{equation}
This update rule corresponds to a special case of the gradient descent method with an adaptive learning rate. Before proving the convergence, we first introduce the following key concepts.

{\bf{Definition 1 \cite{A23}}}: If a function $G_1(x,x^t) \geq G_2(x)$ and $G_1(x,x)=G_2(x)$, then $G_1(x,x^t)$ is called an auxiliary function of $G_2(x)$.

{\bf{Lemma 1}}:If $G_1(x,x^t)$ is an auxiliary function for $G_2(x)$, then the function $G_2(x)$ is non-increasing when the following conditions holds:
\begin{equation}
    {x^t}^{ + 1}\; = {\rm{ }}\mathop {arg\;min\;{G_1}(x,{x^t})}\limits_x 
\end{equation}
Next, let $F_{ij}$ represent ${\Theta _{ij}}$, and we proceed to compute the first and second partial derivatives of $F_{ij}$:
\begin{equation}
    \begin{array}{l}
F_{ij}^{'} = {\left( {2{Y^T}YP - 2{Y^T}B} \right)_{ij}}, 
F_{ij}^{''} = {\left( {2{Y^T}Y} \right)_{ij}}.
\end{array}
\end{equation}
We then perform the Taylor expansion of $F_{ij}$ around the current point:
\begin{equation}
    {F_{ij}}({P_{ij}}) = {F_{ij}}(P_{ij}^t) + F_{ij}^{'}(P_{ij}^t)\left( {{P_{ij}} - P_{ij}^t} \right) + \frac{1}{2}F_{ij}^{''}(P_{ij}^t){\left( {{P_{ij}} - P_{ij}^t} \right)^2}.
\end{equation}
Next, we define the auxiliary function for $F_{ij}(P_{ij})$ as :
\begin{equation}
    G({P_{ij}},P_{ij}^t) = {F_{ij}}(P_{ij}^t) + F_{ij}^{'}(P_{ij}^t)\left( {{P_{ij}} - P_{ij}^t} \right) + \frac{{{{\left( {{Y^T}YP} \right)}_{ij}}}}{{P_{ij}^t}}{\left( {{P_{ij}} - P_{ij}^t} \right)^2}
\end{equation}
where $P_{ij}^t$ represents the previously updated value.

{\bf{Proof}}: If ${P_{ij}} = P_{ij}^t$, then in Eq.(33) we have
$G({P_{ij}},P_{ij}^t) = {F_{ij}}(P_{ij}^t)$.For the other condition $G({P_{ij}},P_{ij}^t) \ge {F_{ij}}({P_{ij}})$ in Definition 1, we need to prove the following inequality:
\begin{equation}
    {\left( {{Y^T}YP} \right)_{ij}} \ge {\left( {{Y^T}Y} \right)_{ij}}{P_{ij}}.
\end{equation}
Clearly, it can be observed that:
\begin{equation}
    {\left( {{Y^T}YP} \right)_{ij}} = \sum\nolimits_{q = 1}^c {{{\left( {{Y^T}Y} \right)}_{iq}}} {P_{qj}} \ge  + {\left( {{Y^T}Y} \right)_{ij}}{P_{ij}}.
\end{equation}
Therefore, Eq.(34) holds, which implies $G({P_{ij}},P_{ij}^t) \ge {F_{ij}}({P_{ij}})$ , and $G({P_{ij}},P_{ij}^t)$ serves as the auxiliary function for ${F_{ij}}({P_{ij}})$ . By substituting auxiliary function into Eq.(30) , we derive the update rule for $P$ based on $\frac{{\partial G({P_{ij}},P_{ij}^t)}}{{\partial {P_{ij}}}} = 0$ as follows:
\begin{equation}
    P_{ij}^{t + 1} \leftarrow P_{ij}^t - P_{ij}^t\frac{{F_{ij}^{'}(P_{ij}^t)}}{{2{{\left( {{Y^T}{P^t}} \right)}_{ij}}}} \leftarrow P_{ij}^t\frac{{{{\left( {{Y^T}B} \right)}_{ij}}}}{{{{\left( {{Y^T}Y{P^t}} \right)}_{ij}}}}.
\end{equation}
Based on the above proof, the objective function is guaranteed to be non-increasing under the conditions specified in Eq.(36). The proof for variable $W$ follows a process similar to that for variable $P$. Therefore, the objective function is non-increasing with respect to both $W$ and $P$.

Regarding the discrete optimization rule for B, it has been demonstrated in the literature \cite{A24} that discrete optimization can effectively converge. Therefore, the convergence of the objective function with respect to the variables $W$, $P$ and B is established. In the subsequent experimental section, we will also provide convergence curves to further illustrate the behavior of the objective function.

\section{Experiments}
In this section, we conduct several experiments to verify the effectiveness of the proposed BHDG method. We evaluate its performance using multiple benchmark datasets across different domains.
\subsection{Datasets}
To evaluate the effectiveness of the proposed BHDG, we conducted comparative experiments on ten benchmark multi-label datasets. These datasets are publicly available for download from the Mulan \cite{A25} website and the UCI \cite{A26} repository. They span various fields, including biology, text, and image domains. Table 2 provides a summary of the characteristics of these datasets. For each dataset, we split the data into training and testing sets. Specifically: training sets for Science, Entertainment, Corel5k, and Yelp comprise 40\% of the total sample size. The training sets for the remaining datasets comprise 50\% of the total sample size. These datasets provide a diverse set of tasks that allow us to comprehensively evaluate the performance of the BHDG method across different domains.
\subsection{Experimental setup}
To evaluate the performance of the proposed BHDG method, We compared it with ten multi-label feature selection methods, including one filter-based method, RF-ML\cite{A27}, and nine sparse regression-based methods: RFS\cite{A28}, ls-l21\cite{A29}, CSFS\cite{A30}, MSSL\cite{A31}, MDFS\cite{A32}, SSFS\cite{A33}, RGFS\cite{A34}, DRMFS\cite{A35}, and MSFS\cite{A15}. 

{\bf{Parameter Settings}}: Optimal parameter values for all baseline methods were selected based on the parameter sensitivity experiments reported in the corresponding literature. For the proposed BHDG, the dimension of the binary hashing label $l$ was set to half the number of labels, and parameters $\lambda_1$,$\lambda_2$ and $\lambda_3$ were tuned over $\{10^{-2}, 10^{-1},..., 10^{3}\}$, while $\alpha$ was chosen from $\{0.1, 0.3, 0.5, 0.7, 0.9, 1\}$, and $\rho$ was varied over $\{0.01, 0.05, 0.1, 0.15, 0.2, 0.25\}$.

{\bf{Classification Metrics}}: For classification performance comparison, we employed the ML-KNN \cite{A36} classifier compute six evaluation metrics for classification comparison experiments, including Hamming Loss (HL), Ranking Loss (RL), One Error (OE), Coverage (CV), Average Precision (AP), and Macro-$F_1$ \cite{A37}. The number of neighbors in the ML-KNN classifier was set to 10, and the smoothing parameter was set to 1. 

{\bf{Additional Settings}}: We set $\sigma$ to 1.0 for all kernel functions, and the number of neighbors was set to 10 for constructing similarity matrices $S_X$, $S_Y$ and $S_B$ . 

{\bf{Computational Environment}}: All experiments were conducted in a Python environment on a PC equipped with an Intel Core i5-14400 processor (2.50 GHz) and 32 GB RAM.


\begin{table*}
\renewcommand\arraystretch{1.3}
    \caption{The descriptions of datasets}\centering
    \resizebox{0.8\linewidth}{!}{
    \begin{tabular}{lllllll}
    \hline
        Datasets & Instances &Training  &Testing  &Features  &Labels  &Domin \\
         \hline
         Amphibians&	189&95	&94  & 		23&	7&	Biology        \\
       LangLog &1460	&730&730  &		1004&	75&	Text          \\
       Enron&	1702&851	&851  &		1001&	53&	Text          \\
      reuters&	2000&	1000&	1000&	243&	7&	Text              \\
       image&	2000&	1000&	1000&	294&	5&	Image            \\
        Yeast&	2417&	1209&	1208&	103&	14&	Biology            \\ 
        Science&5000&2000&3000&743&40&Text\\ 
        Entertainment&5000&2000&3000&640&21&Text\\ 
        Corel5k&	5000&	2000&	3000&	499&	374&	Image\\
        Yelp&10810 &4324	&6486	&	671&	5& Text\\
        \hline
    \end{tabular}}
    \label{tab:my_label}
\end{table*}
\subsection{Results and Discussion}
The classification results and analysis for the proposed method, BHDG, are summarized in this subsection. Tables 3-8 record the experimental results across six evaluation metrics (HL, RL, OE, CV, AP, Macro-$F_1$) for the top 20\% of selected features. In these tables, an upward arrow ($\uparrow$) denotes metrics where higher values are better, while a downward arrow ($\downarrow$) signifies metrics where lower values are preferable. The best-performing method for each dataset is highlighted in bold font. The following observations can be drawn from Tables 3-8:

\begin{itemize}
    \item Table 3 (HL): BHDG achieved the highest ranking on all 10 datasets for this metric, followed by SSFS, MDFS, ls-l21, DRMFS, RF-ML, CSFS, MSFS, RFS, MSSL, and RGFS. 
    \item Table 4 (RL): BHDG performed best across all datasets for this metric, followed by SSFS, ls-l21, MDFS, RFS, DRMFS, RGFS, MSFS, CSFS, RF-ML, and MSSL 
    \item Table 5 (OE): BHDG ranked the highest for this metric, followed by RFS, ls-l21, MSFS, SSFS, RF-ML, DRMFS, MDFS, MSSL, CSFS, and RGFS. BHDG secured the best results on 9 datasets.
    \item Table 6 (CV): BHDG achieved the top ranking for this metric, followed by ls-l21, SSFS, MDFS, RFS, DRMFS, MSFS, RGFS, CSFS, RF-ML, and MSSL. BHDG excelled on 8 datasets.
    \item Table 7 (AP): BHDG achieved the top ranking across all datasets for this metric, followed by ls-l21, SSFS, MDFS, DRMFS, RFS, MSFS, RGFS, CSFS, RF-ML, and MSSL.
    \item Table 8 (Macro-$F_1$): BHDG achieved the highest ranking for this metric, followed by ls-l21, SSFS, DRMFS, MDFS, RGFS, RFS, RF-ML, CSFS, MSSL, and MSFS. BHDG led on 9 datasets.
    
\end{itemize}

\begin{table*}

\begin{flushleft}
\renewcommand\arraystretch{1.3}
    \caption{Results of various feature selection algorithms with respect to HL($\downarrow$ )}\centering
   \fontsize{10}{12}\selectfont
   \resizebox{\linewidth}{!}{
    \begin{tabularx}{1.15\linewidth}{llllllllllll}
    \hline
        Datasets &RFS&ls-l21&RF-ML&CSFS&MSSL&MDFS&SSFS&RGFS&DRMFS &MSFS&BHDG \\
         \hline
        Amphibians&0.3326&0.3231&0.3215&0.3289&0.3374&0.3325&0.3215&0.3398&0.3293&0.3316&\textbf{0.318}\\
        LangLog &0.0158&0.0158&0.0159&0.0159&0.0159	&0.0159&0.0159&	0.0159&0.0158&0.0159&\textbf{0.0157}\\
        Enron &0.0547&0.053&0.0585&0.0567&0.0536&0.0557&	0.052&0.0522&0.0536&0.0559&\textbf{0.0517}\\
        reuters &0.123&0.123&0.108&0.1233&0.109&0.122&	0.106&0.118&0.109&0.1233&\textbf{0.0999}\\
        image &0.208&0.203&0.2148&0.214&0.212&0.2017&0.2061	&0.2028&0.2061&0.2086&\textbf{0.1976}\\
        Yeast &0.2105&0.2101&0.2164&0.2105&0.2105&0.2082&	0.2103&0.2107&0.2186&0.2105&\textbf{0.204}\\
        Science&0.0354&	0.0353&	0.0356&	0.0354&	0.0355&	0.0348&	0.0352&	0.0354&	0.0348&	0.0352&	\textbf{0.0344}
\\
        Entertainment&0.0605&	0.0643&	0.0638&	0.0649&	0.0637&	0.0652&	0.0665&	0.0653&	0.061&	0.0611&	\textbf{0.0603}
\\
        Corel5k &0.00941&0.00943&0.00945&0.00945&0.00946&0.0095&0.00943		&0.00943&0.00945&0.00943&\textbf{0.0094}\\
        Yelp &0.1721&0.1728&0.1753&0.1747	&0.1796&0.1747&0.1732&0.1786&0.171&0.1759&\textbf{0.1693}\\
         \hline
         Average rank&  7.05& 5.4&  8.2&  8.1&  7.5&  4.85& 4.15& 6.75& 5.4&  7.6&  \textbf{1.} \\
         \hline
    \end{tabularx}}
    \end{flushleft}
    \label{tab:my_label}
\end{table*}

\begin{table*}
\renewcommand\arraystretch{1.3}
    \caption{Results of various feature selection algorithms with respect to RL($\downarrow$ )}\centering
   \fontsize{10}{12}\selectfont
   \resizebox{\linewidth}{!}{
    \begin{tabular*}{1.15\linewidth}{llllllllllll}
    \hline
        Datasets &RFS&ls-l21&RF-ML&CSFS&MSSL&MDFS&SSFS&RGFS&DRMFS &MSFS&BHDG \\
         \hline
        Amphibians &0.4354&0.4281&0.4263&0.427&0.4472&0.4377&	0.4163&0.4525&0.4362&0.4334&\textbf{0.4067}\\
        LangLog &0.8452&0.8527&0.8565&0.8472&0.8575&0.8503&	0.8561&	0.8544&0.8571&0.8541&\textbf{0.8251}\\
        Enron &0.6995&0.6772&0.7948&0.7546&0.6903&0.7576&0.6509		&0.6684&0.6954&0.725&\textbf{0.6238}\\
        reuters &0.5995&0.6072&0.5065&0.607&0.5336&0.5889&0.5073&0.5761&0.5308&0.5889&\textbf{0.4733}\\
        image &0.672&0.6654&0.7502&0.7383&0.7331&0.6675&0.6624&0.693&0.6919&0.6892&\textbf{0.6582}\\
        Yeast &0.5173&0.5144&0.5673&0.5181&0.5337&0.5102&0.5132		&0.5261&0.581&0.5251&\textbf{0.4914}\\
        Science& 0.9722&	0.9647&	0.9947&	0.9666&	0.972&	0.9309	&0.9527	&0.9543&	0.9621&	0.9605&	\textbf{0.9277}
\\
        Entertainment&0.8247&	0.9189&	0.9137&	0.9376&	0.9119&	0.9375&	0.9705&	0.9418&	0.8503&	0.8419&	\textbf{0.8414}
\\
        Corel5k &0.9944&0.994&0.9943&0.9952&0.9977&0.9954&0.9972&0.9972&0.9946&0.9955&\textbf{0.9931}\\
        Yelp &0.6431&0.6377&0.6648&0.6874&0.693&0.6778&0.6615&0.7039	&0.642&0.6956&\textbf{0.6087}\\
         \hline
         Average rank&  6.&   4.9&  7.7&  7.2&  8.7&  5.7&  4.85& 6.8&  6.15& 7.&   \textbf{1.} \\
         \hline
    \end{tabular*}}
    \label{tab:my_label}
\end{table*}

\begin{table*}
\renewcommand\arraystretch{1.3}
    \caption{Results of various feature selection algorithms with respect to OE($\downarrow$ )}\centering
   \fontsize{10}{12}\selectfont
   \resizebox{\linewidth}{!}{
    \begin{tabular*}{1.15\linewidth}{llllllllllll}
    \hline
        Datasets &RFS&ls-l21&RF-ML&CSFS&MSSL&MDFS&SSFS&RGFS&DRMFS&MSFS&BHDG \\
         \hline
        Amphibians &0.371&0.3666&0.3122&0.3394&\textbf{0.3114}&\textbf{0.3114}&\textbf{0.3114}&0.3122&0.3166&\textbf{0.3114}&0.3415\\
        LangLog &0.01&0.0118&0.0144&0.0115&0.0129&0.0145&0.0142&0.0112&0.0116&0.0137&\textbf{0.0085}\\
        Enron &0.0165&0.0159&0.0158	&0.0159&0.0165&0.0166&0.016	&0.0158&0.0155&0.0159&\textbf{0.0153}\\
        reuters &0.4938	&0.5007&0.421&0.4961&0.4483&0.4905&0.4222&0.4768	&0.4375&0.4916&\textbf{0.4021}\\
        image &0.1613&0.1461&0.1667&0.1605&0.1624	&0.1526&0.152&0.1698&0.1504&0.1551&\textbf{0.1434}\\
        Yeast &0.2691&0.2772&0.3104&0.2854&0.2819&0.2641&0.28576		&0.2888&0.3153&0.2795&\textbf{0.2548}\\
        Science&0.9333&	0.929&	0.96&	0.925&	0.934&	0.9077&	0.9117&	0.91&	0.9235&	0.919&	\textbf{0.8843}
\\
        Entertainment&0.7924&	0.8957&	0.8897&	0.9173&	0.885&	0.9137&	0.95&	0.92&	0.8187&	0.8103&	\textbf{0.8087}
\\
        Corel5k &0.9709&0.9718&0.9682&0.9711&0.9784&0.9705&0.9774&0.9774&0.9694&0.9729&\textbf{0.9673}\\
        Yelp &0.8244&0.8186&0.8429&0.8661&0.8663&0.8595&0.8407&0.888&0.8215&0.875&\textbf{0.7965}\\
         \hline
         Average rank&  6.75& 6.05& 7.15& 6.85& 8.2&  6.1&  5.7&  6.&   4.3&  7.3&  \textbf{1.6} \\
         \hline
    \end{tabular*}}
    \label{tab:my_label}
\end{table*}

\begin{table*}
\renewcommand\arraystretch{1.3}
    \caption{Results of various feature selection algorithms with respect to CV($\downarrow$ )}\centering
   \fontsize{10}{12}\selectfont
   \resizebox{\linewidth}{!}{
    \begin{tabular*}{1.15\linewidth}{llllllllllll}
    \hline
        Datasets &RFS&ls-l21&RF-ML&CSFS&MSSL&MDFS&SSFS&RGFS&DRMFS &MSFS&BHDG \\
         \hline
        Amphibians &4.086&4.042&\textbf{4.012}&4.028&4.13&4.0806&4.0122		&4.1368&4.1307&4.0675&4.023\\
        LangLog &73.974&73.982&73.987&73.976&73.988&73.98&73.986		&73.984&73.987&73.984&\textbf{73.953}\\
        Enron &51.16&51.115&51.37&51.283&51.082&51.255&51.053&51.061	&51.147&51.199&\textbf{51.002}\\
        reuters &2.508&2.532&2.02&2.535&2.156&2.46&2.02	&2.329&2.169&2.463&\textbf{1.89}\\
        image &3.5056&3.5286&3.5696&3.5668&3.552&3.5245&3.5301&3.5031&3.5403&3.5&\textbf{3.495}\\
        Yeast &11.052&11.064&11.067&11.071&11.054&11.052&11.0597&11.084&11.07&11.042&\textbf{11.031}\\
        Science&24.526&	24.362&	25.122&	24.436&	24.725&	23.927&	24.208&	24.313&	24.434&	24.382&	\textbf{23.542}
\\
        Entertainment&6.401&	6.81&	6.718&	6.95&	6.734&	6.903&	7.238&	6.92&	6.532&	\textbf{6.442}&	6.492
\\
        Corel5k &207.81&207.69&207.9&207.79&207.98&207.92&208.03&208.03&207.67&207.83&\textbf{207.36}\\
        Yelp &0.9016&0.8749&0.9399&0.9292&0.9657&0.9047&0.9542&1.013&0.8215&0.9348&\textbf{0.8261}\\
         \hline
         Average rank&  5.9&  5.4&  7.95& 7.4&  8.15& 5.6&  5.4&  6.75& 6.05& 6.1&  \textbf{1.3} \\
         \hline
    \end{tabular*}}
    \label{tab:my_label}
\end{table*}

\begin{table*}
\renewcommand\arraystretch{1.3}
    \caption{Results of various feature selection algorithms with respect to AP($\uparrow$ )}\centering
   \fontsize{10}{12}\selectfont
   \resizebox{\linewidth}{!}{
    \begin{tabular*}{1.15\linewidth}{llllllllllll}
    \hline
        Datasets &RFS&ls-l21&RF-ML&CSFS&MSSL&MDFS&SSFS&RGFS&DRMFS &MSFS&BHDG \\
         \hline
        Amphibians &0.4549&0.4618&0.4689&0.4612&0.45766&0.4606&	0.4689&0.4622	&0.4663&0.4626&\textbf{0.4759}\\
        LangLog &0.9843&0.9842&0.9842&0.9843&0.9841&0.9842&	0.9842&0.9842&0.9842&0.9842&\textbf{0.9846}\\
        Enron &0.9442&0.9451&0.94&0.9416&0.9453&0.9424&	0.9466&0.9459	&0.9449&0.9425&\textbf{0.9473}\\
        reuters &0.2726&0.2702&0.3235&0.2716&0.3114&0.2793&	0.3193&0.2716	&0.313&0.2743&\textbf{0.3393}\\
        image &0.8056&0.8087&0.7932&0.7955&0.7971&0.8086&	0.8039&0.8055&0.8034&0.8075&\textbf{0.8111}\\
        Yeast &0.7204&0.7212&0.7135&0.7203&0.7191&0.7203&0.722		&0.7199&0.7103&0.7193&\textbf{0.7263}\\
        Science&0.0411&	0.0444&	0.0366&	0.0437&	0.0384&	0.0488&	0.0425&	0.0442&	0.042&	0.0452&	\textbf{0.0526}
\\
        Entertainment&0.1331&	0.114&	0.1197&	0.1018&	0.1157&	0.1035&	0.0803&	0.0994&	0.1202&	0.1183&	\textbf{0.1271}
\\
        Corel5k &0.01	&0.01&0.0099&0.0098	&0.0097	&0.0099&0.0098&0.0098&0.01&0.01&\textbf{0.0101}\\
        Yelp &0.249	&0.2607&0.2333&0.2322&0.2084&0.2416&0.2383		&0.2177&0.2618&0.2275&\textbf{0.279}\\
         \hline
         Average rank  &6.2&  4.7&  7.95& 7.7&  8.9&  5.7&  5.&   6.65& 5.9 & 6.3&  \textbf{1.} \\
         \hline
    \end{tabular*}}
    \label{tab:my_label}
\end{table*}

\begin{table*}
\renewcommand\arraystretch{1.3}
    \caption{Results of various feature selection algorithms with respect to Macro-$F_1$($\uparrow$ )}\centering
   \fontsize{10}{12}\selectfont
   \resizebox{\linewidth}{!}{
    \begin{tabular*}{1.15\linewidth}{llllllllllll}
    \hline
        Datasets &RFS&ls-l21&RF-ML&CSFS&MSSL&MDFS&SSFS&RGFS&DRMFS &MSFS&BHDG \\
         \hline
        Amphibians &0.6&0.6147&0.6291&0.6119&0.5995&0.6051&	0.6291&0.5959&0.6085&0.6104&\textbf{0.6334}\\
        LangLog &0.992&\textbf{0.9921}&0.992&0.992&0.992&0.992&0.9919&\textbf{0.9921}&0.992&0.9919&\textbf{0.9921}\\
        Enron &0.9713&0.9722&0.9694&0.9703&0.9718&0.9708&0.9727&0.9707	&0.9718&0.9707&\textbf{0.9728}\\
        reuters &0.5022&0.496&0.5948&0.4971&0.571&0.5135&0.5911&0.5192&0.5711&0.5109&\textbf{0.6227}\\
        image &0.8725&0.8751&0.8711&0.871&0.8723&0.8769&0.8749	&0.877&0.875&0.8718&\textbf{0.8793}\\
        Yeast &0.8578&0.8579&0.8563&0.858&0.859&0.8603&	0.8575&0.8582	&0.8555&0.8584&\textbf{0.8634}\\
        Science&0.0547&	0.0678&	0.0093&	0.0651&	0.0453&	0.1174&	0.0745&	0.0789&	0.0596&	0.0725&	\textbf{0.1239}
\\
        Entertainment&0.2544&	0.1357&	0.1462&	0.0928&	0.1449&	0.1109&	0.058&	0.1014&	0.2221&	0.2461&	\textbf{0.2463}
\\
        Corel5k &0.0113&0.016&\textbf{0.0114}&0.0098&0.0046&0.0094&0.0055		&0.0055&0.0164&0.0091&0.0104\\
        Yelp &0.2374&0.2484&0.1864&0.168&0.1477&0.1896&	0.1915&0.1148&0.2462&0.1498&\textbf{0.2914}\\
         \hline
         Average rank  &6.95& 4.6&  7.4&  7.75& 7.8&  5.45& 5.15& 6.5&  5.2&  7.8&  {\bf{1.4}} \\
         \hline
    \end{tabular*}}
    \label{tab:my_label}
\end{table*}

From the results of Tables 3-8, BHDG achieved the best average ranking across all evaluation metrics, surpassing the second-best method by at least 2.7 ranks. These findings demonstrate that the innovations proposed in this study significantly improved feature weight quality, particularly through the pursuit of more reliable latent representation labels.

\begin{figure*}
    \centering
    \includegraphics[width=1\linewidth]{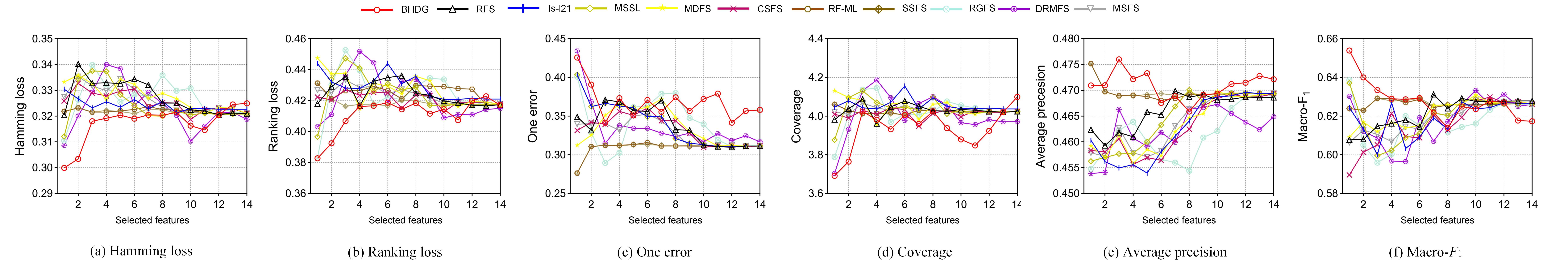}
    \caption{Results on Amphibians with different numbers of features.}
    \label{fig:enter-label}
\end{figure*}

\begin{figure*}
    \centering
    \includegraphics[width=1\linewidth]{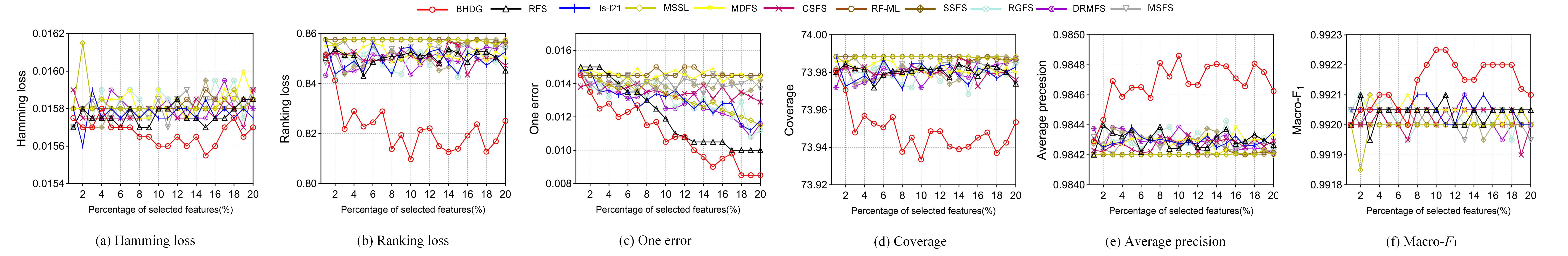}
    \caption{Results on Enron with different numbers of features.}
    \label{fig:enter-label}
\end{figure*}

\begin{figure*}
    \centering
    \includegraphics[width=1\linewidth]{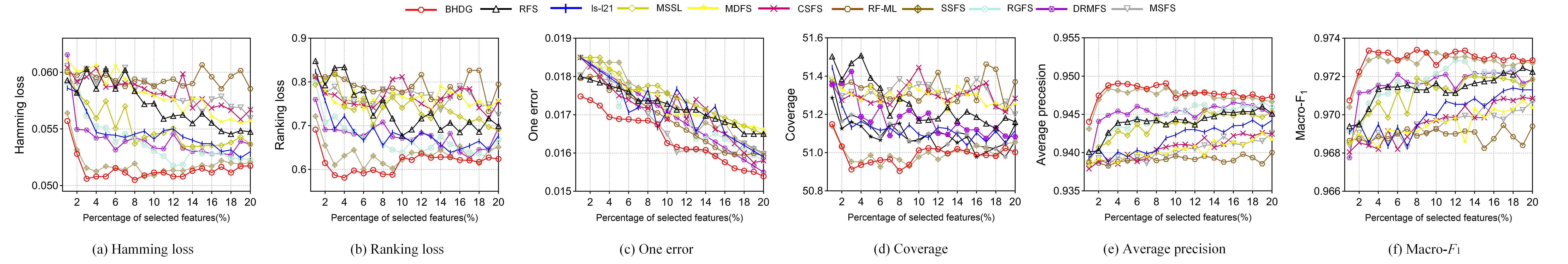}
    \caption{Results on Yeast with different numbers of features.}
    \label{fig:enter-label}
\end{figure*}

\begin{figure*}
    \centering
    \includegraphics[width=1\linewidth]{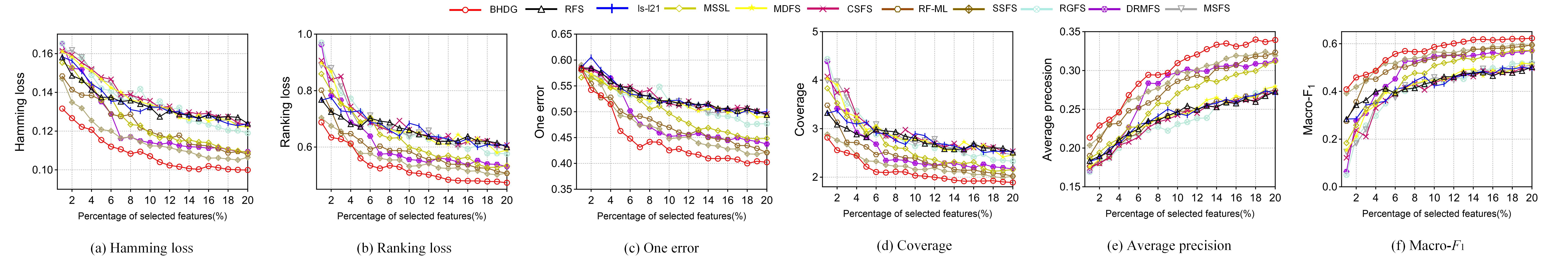}
    \caption{Results on Reuters with different numbers of features.}
    \label{fig:enter-label}
\end{figure*}

\begin{figure*}
    \centering
    \includegraphics[width=1\linewidth]{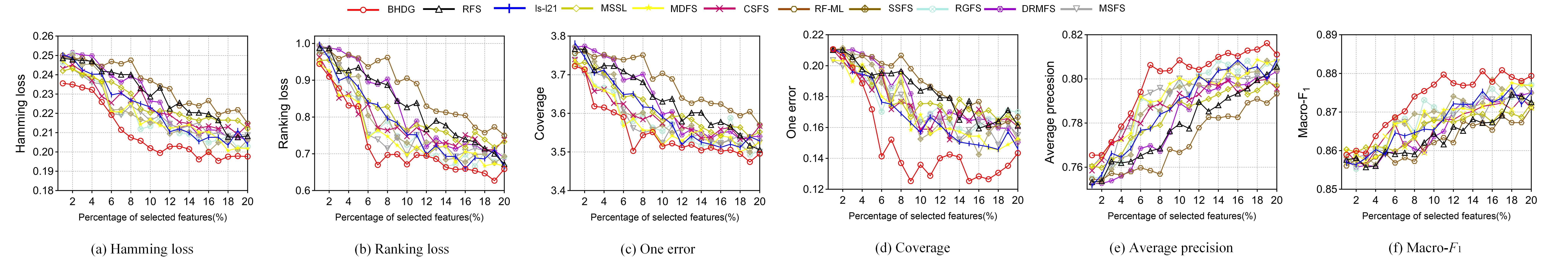}
    \caption{Results on image with different numbers of features.}
    \label{fig:enter-label}
\end{figure*}

\begin{figure*}
    \centering
    \includegraphics[width=1\linewidth]{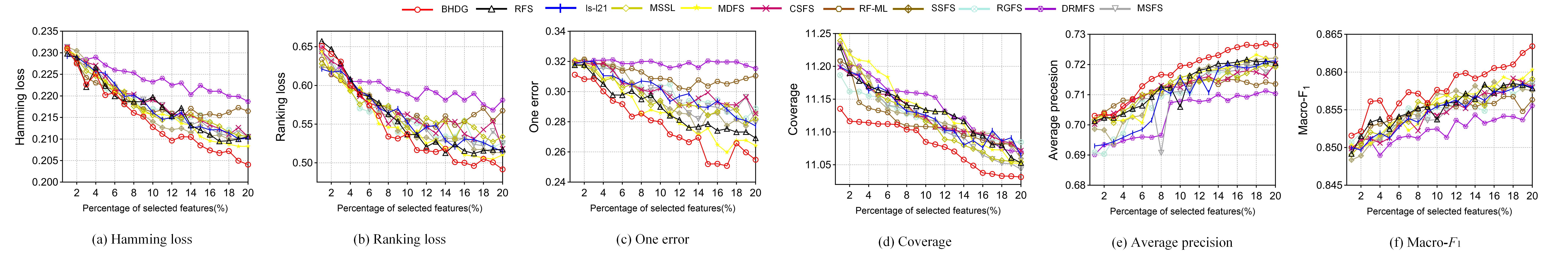}
    \caption{Results on Langlog with different numbers of features.}
    \label{fig:enter-label}
\end{figure*}

\begin{figure*}
    \centering
    \includegraphics[width=1\linewidth]{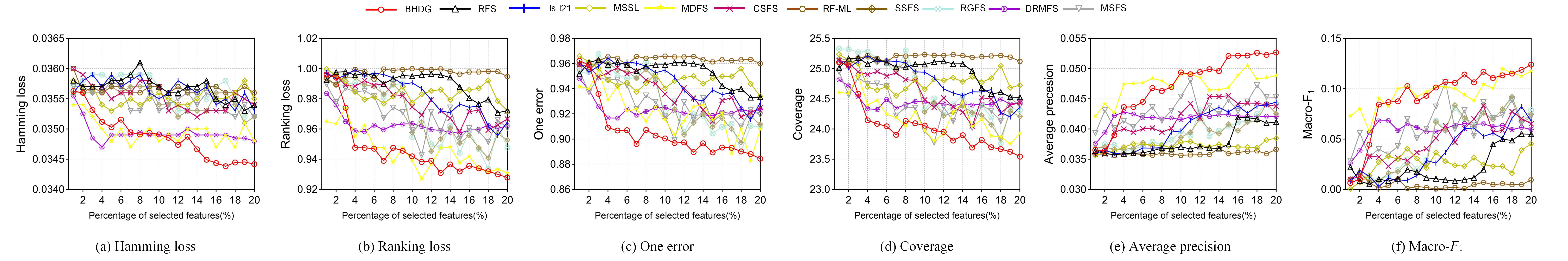}
    \caption{Results on Science with different numbers of features.}
    \label{fig:enter-label}
\end{figure*}

\begin{figure*}
    \centering
    \includegraphics[width=1\linewidth]{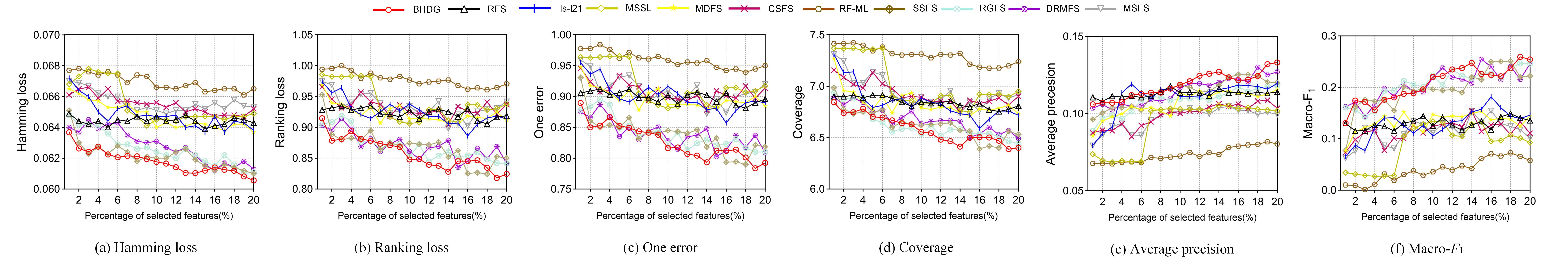}
    \caption{Results on Entertainment with different numbers of features.}
    \label{fig:enter-label}
\end{figure*}

\begin{figure*}
    \centering
    \includegraphics[width=1\linewidth]{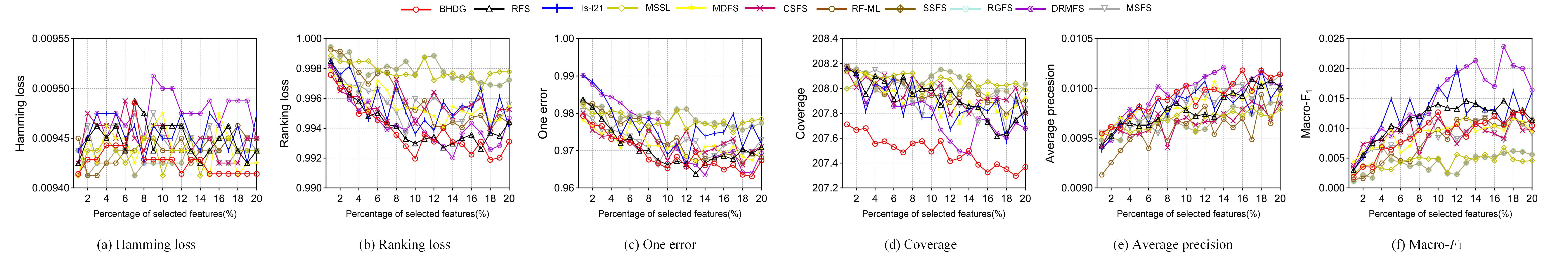}
    \caption{Results on Corel5k with different numbers of features.}
    \label{fig:enter-label}
\end{figure*}

\begin{figure*}
    \centering
    \includegraphics[width=1\linewidth]{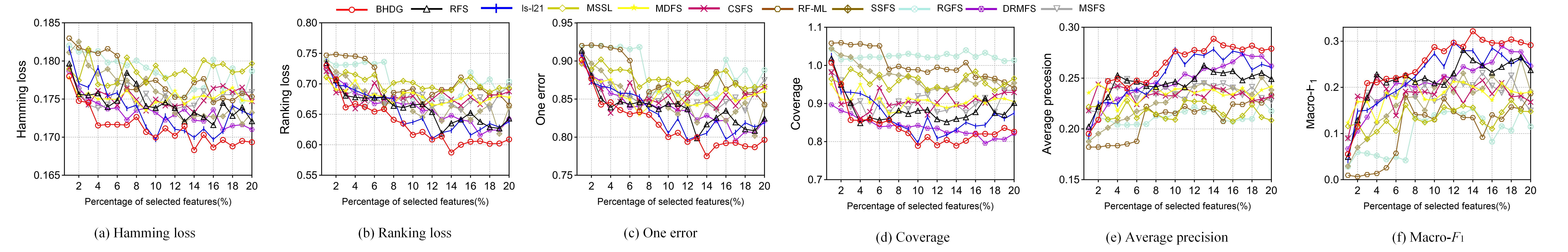}
    \caption{Results on yelp with different numbers of features.}
    \label{fig:enter-label}
\end{figure*}

{\bf{Visual Analysis (Figs. 3–12)}}: Figs.3-12 illustrate the performance variations across 10 datasets when selecting different numbers of features. For all datasets except Amphibians, the selected features range from 2\% to 20\% in increments of 1\%. Due to the smaller feature count in Amphibians, the selected features range from 2 to 14. The visual results reveal that, apart from the Amphibians dataset, BHDG outperformed other methods in most cases. On the Amphibians dataset, however, BHDG demonstrated a significant advantage only when selecting 2-5 features. In this experiment, we found that ls-l21, MDFS, and SSFS were the top-performing comparison methods. Among these, DFS incorporates two general graph constraints, while SSFS applies a single graph constraint. The findings suggest that multi-graph constraints are more effective at extracting information than single-graph constraints. Nevertheless, SSFS uses a distinct shared semantic regression model, which also contributes to its competitive performance. Despite this, both MDFS and SSFS employ fixed graph constraints and do not consider binary pseudo-label learning, which ultimately results in lower classification performance compared to BHDG. Similarly, ls-l21, which adopts a simpler model to learn feature weights, achieved competitive results in the experiments. This can be attributed to its use of a smooth convex optimization approach to solve the objective function, which accelerates the computation process and enhances algorithm performance.

\begin{table}
\renewcommand\arraystretch{1.3}
    \centering
    \caption{Results of Friedman’s statistical test compared to 11 methods at the significance level $\alpha = 0.05$.}
    \resizebox{0.5\linewidth}{!}{
    \begin{tabular}{lll}
    \hline
   Evaluation metric	&${F_F}$	&Critical value($\alpha = 0.05$)    \\
    \hline
       HL&6.514  & \multirow{6}*{{\small 1.937} } \\
       RL&5.328& \\    
        OE&3.609& \\
       CV&4.016& \\
        AP &5.874& \\
Macro-$F_{1}$&4.643 & \\
         \hline
    \end{tabular}}
    \label{tab:my_label}
\end{table}
\subsection{Statistical significance in performance comparisons}
To further analyze the relative performance of the proposed method BHDG against other methods, we employed the Friedman test \cite{A38} for significance analysis. Given $K$ algorithms and $N$ data sets, with $R_j$ representing the average ranking of the $j$-th algorithm across all datasets, Friedman's statistical value is calculated as follows:
\begin{equation}
    \begin{array}{l}
x_F^2 = \frac{{12N}}{{K(K + 1)}}\left( {\sum\limits_{i = 1}^k {R_i^2 - \frac{{K{{(K + 1)}^2}}}{4}} } \right),\\
{F_F} = \frac{{(N - 1)x_F^2}}{{N(K - 1) - x_F^2}}.
\end{array}
\end{equation}
We calculated the Friedman's statistical value $F_F$ for each metric using the average rankings in Tables 3-8, as summarized in Table 9. From Table 9. At a significance level of 0.05, the null hypothesis—that all methods perform identically—is clearly rejected for each metric. 

To identify specific performance differences, we conducted a Nemenyi test for post-hoc analysis. If the average ranking difference between the two methods exceeds a critical difference (CD), their performance is considered significantly different. The corresponding CD values
${q_\alpha }\sqrt {\frac{{k\left( {k + 1} \right)}}{{6N}}} $ and ${q_\alpha } = {\rm{3}}{\rm{.525}}$ were determined. The results of the Nemenyi test for BHDG and ten other methods across six evaluation metrics are presented in Fig.13. The horizontal axis represents the ranking order in each subfigure, with higher rankings displayed on the right. Algorithms connected by a black line have average rankings within one CD, indicating no significant performance differences. As shown, the proposed method achieved the highest ranking among all compared methods. The key findings are as follows. BHDG is significantly superior to RF-ML, CSFS, MSSL, and RGFS across all evaluation metrics. BHDG outperforms DRMFS on RL and OE, surpasses MDFS on OE, and is better than the state-of-the-art method MSFS on HL, RL, AP, and Macro-$F_1$. 

In summary, the Friedman test and Nemenyi test validate that the proposed method demonstrates significantly superior classification performance compared to the other methods.

\begin{figure*}
    \centering
    \includegraphics[width=0.8\linewidth]{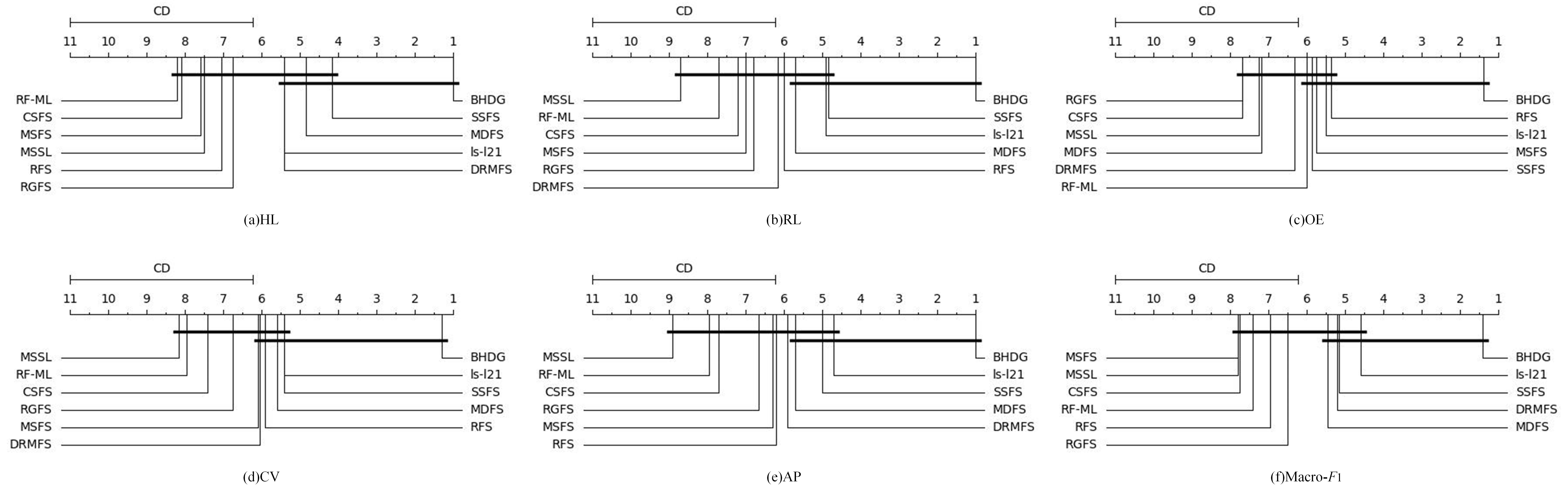}
    \caption{Nemenyi test on 11 algorithms for HL,RL,OE,CV,AP, and Macro-$F_1$(CD= 4.774 at 0.05 significance level).}
    \label{fig:enter-label}
\end{figure*}

\subsection{Parameter sensitivity analysis}
To assess the sensitivity of the proposed method to parameter settings, we evaluated the impact of parameters $\lambda_1$, $\lambda_2$, $\lambda_3$, $\rho$, and $\alpha$ on the performance of the BHDG algorithm. For simplicity, the Reuters dataset was used for these experiments. During the analysis, one parameter was varied independently while keeping all other parameters fixed. The results of these experiments under different parameter settings are depicted in Figs.14-18. The following insights can be derived from the figures:
\begin{itemize}
    \item $\lambda_1$(Fig.14): The algorithm achieves optimal results on HL, RL, CV, and Macro-$F_1$ metrics when $\lambda_1$ is set to 1000, and the best results on OE and AP when $\lambda_1$ is set to 100.
    \item $\lambda_2$(Fig.15): The algorithm performs best on RL and CV when $\lambda_2=1$, and achieves the highest performance on HL, OE,AP, and Macro-$F_1$ when $\lambda_2=10$.
    \item $\lambda_3$(Fig.16): The algorithm performs optimally on all metrics when $\lambda_3=1000$. 
    \item $\rho$(Fig.17): The algorithm achieves the best performance on HL and AP when $\rho=0.01$, on OE and Macro-$F_1$ when $\rho=0.1$, and on RL and CV when $\rho=0.2$.
    \item $\alpha$(Fig.18): The algorithm performs best on RL, CV, and Macro-$F_1$ when $\alpha=0.3$, on HL and AP when $\alpha$ is set to 0.7, and on OE when $\alpha=1$.
\end{itemize}
Overall, the algorithm is more sensitive to changes in $\lambda_1$, $\lambda_2$, and $\lambda_3$, while it is less sensitive to changes in $\rho$ and $\alpha$. Based on these observations, we recommend setting the parameters $\lambda_1$, $\lambda_2$, $\lambda_3$, $\rho$, and $\alpha$ to {1000, 10, 1000, 0.01, 1}.

\begin{figure*}[h]
    \centering
    \includegraphics[width=1\linewidth]{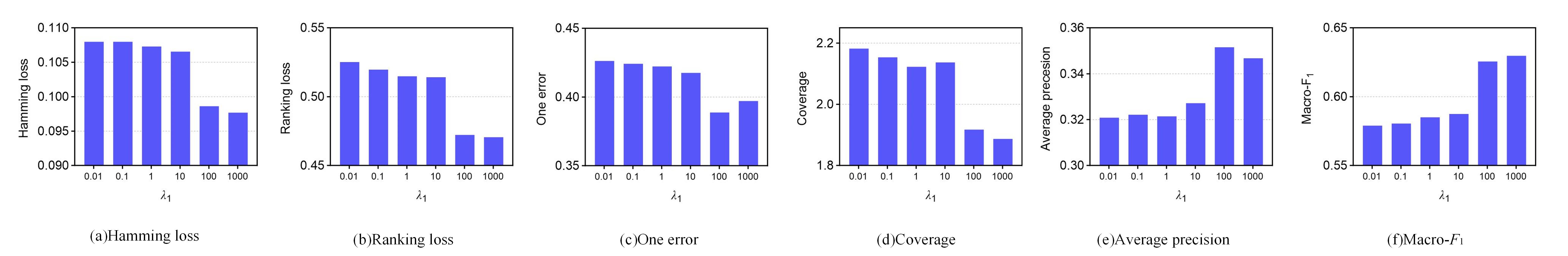}
    \caption{Classification results with different $\lambda_1$ while keeping other parameters unchanged.}
\end{figure*}

\begin{figure*}[h]
    \centering
    \includegraphics[width=1\linewidth]{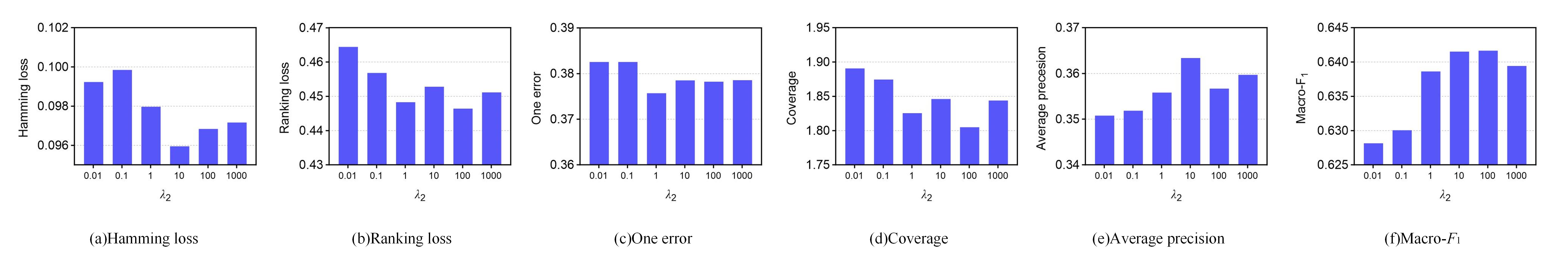}
    \caption{Classification results with different $\lambda_2$ while keeping other parameters unchanged.}
\end{figure*}

\begin{figure*}[h]
    \centering
    \includegraphics[width=1\linewidth]{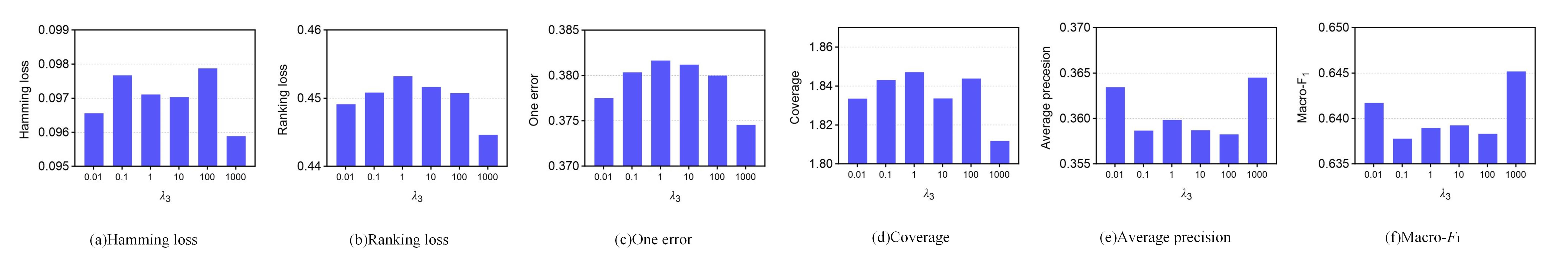}
    \caption{Classification results with different $\lambda_3$ while keeping other parameters unchanged.}
\end{figure*}

\begin{figure*}[h]
    \centering
    \includegraphics[width=1\linewidth]{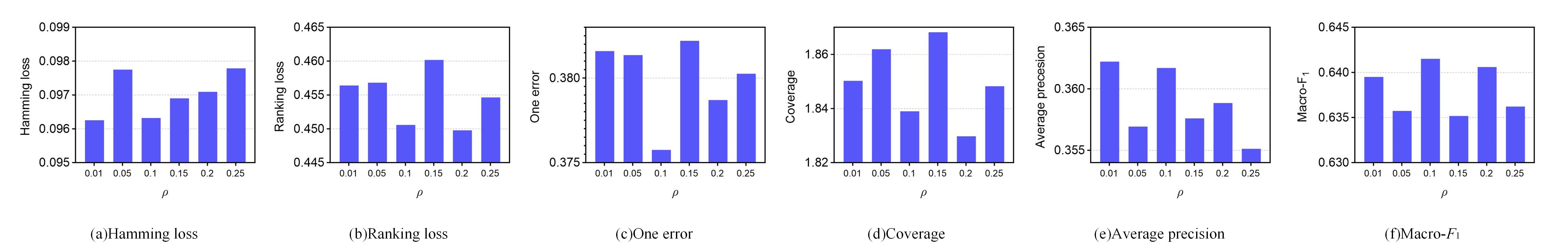}
    \caption{Classification results with different $\rho$ while keeping other parameters unchanged.}
\end{figure*}

\begin{figure*}[!]
    \centering
    \includegraphics[width=1\linewidth]{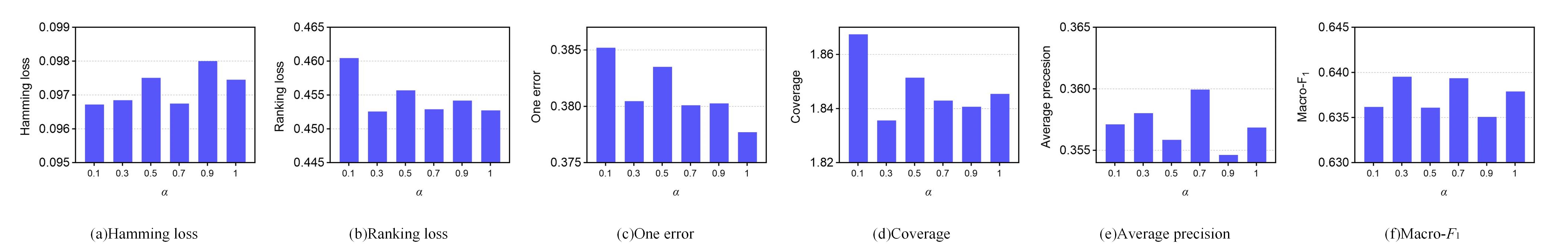}
    \caption{Classification results with different $\alpha$ while keeping other parameters unchanged.}
\end{figure*}

\begin{figure*}
    \centering
    \includegraphics[width=1\linewidth]{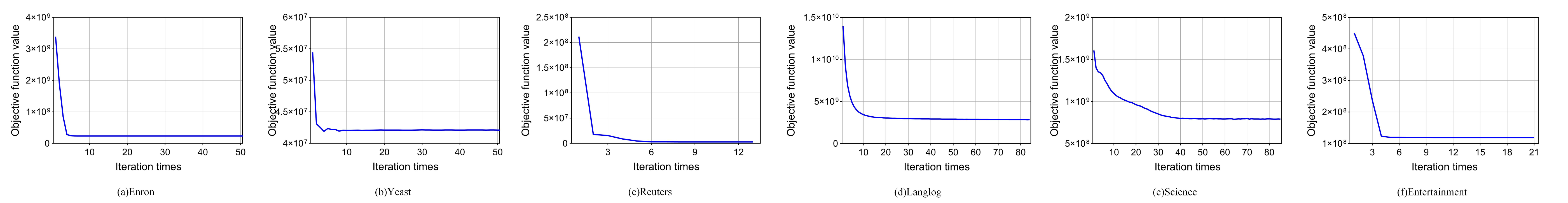}
    \caption{Convergence analysis.}
\end{figure*}

\begin{table}
\renewcommand\arraystretch{1.3}
    \centering
    \caption{Results of the ablation study of BHDG on four datasets.}
    \fontsize{10}{12}\selectfont
    \resizebox{0.7\linewidth}{!}{
    \begin{tabular}{lllllllll}
    \hline
         Dataset&Algorithm  &HL  &RL  &OE  &CV  &AP  &Macro-${F_1}$&Win \\
         \hline
         &BHDG  &0.0524&	{\bf{0.6233}}&	{\bf{0.0151}}&	{\bf{51}}&	0.9471&	{\bf{0.9723}}&{\bf{4}}\\
        Enron &BHDG1  &0.0524&	0.6249&	0.0152&	51.15&	0.9471&	0.9718&0\\
         &BHDG2  &{\bf{0.0521}}&	0.6278&	0.0152&	51.05&{\bf{0.9472}}&
	0.972&{\bf{2}}\\
         \hline
         &BHDG  &{\bf{0.0971}}&	0.4654&	0.389&	1.889&	{\bf{0.3584}}&	{\bf{0.6326}}&{\bf{3}}\\
        reuters &BHDG1  &0.0971&	0.4728&	0.3963&	1.882&	0.3457&	0.6289&0\\
         &BHDG2  &0.0981&	{\bf{0.4589}}&	{\bf{0.384}}&	{\bf{1.865}}&	0.3538&	0.6302&{\bf{3}}\\
         \hline
         &BHDG&  {\bf{0.2012}}&	{\bf{0.6604}}&	{\bf{0.1327}}&	{\bf{3.511}}&	{\bf{0.8108}}&	{\bf{0.8771}}&{\bf{6}}\\
        image &BHDG1  &0.2089&	0.6963&	0.144&	3.538&	0.8031&	0.8731&0 \\
         &BHDG2  &0.2027&	0.669&	0.1343&	3.519&	0.8091&	0.8763&0\\
         \hline
         &BHDG  &{\bf{0.211}}&	{\bf{0.5217}}&	0.2772&	{\bf{11.04}}&	{\bf{0.7191}}&	{\bf{0.8577}}&{\bf{5}}\\
        Yeast &BHDG1  &0.2131&	0.5248&	{\bf{0.2741}}&	11.08&	0.7179&	0.8562 &1\\
         &BHDG2  &0.2147&	0.5377&	0.283&	11.09&	0.7159&	0.8558&0\\
         \hline
    \end{tabular}}
    
    \label{tab:my_label}
\end{table}

\subsection{ Ablation study}
To evaluate the effectiveness of each component in BHDG, we conducted an ablation study by comparing its classification performance on four datasets (Enron, Reuters, Image, Yeast) under the following two modified scenarios:
\begin{itemize}
    \item {\bf{BHDG1:}} Set $\lambda_2$=0, meaning that the dynamic graph constraint term is removed from the objective function of BHDG.

    \item {\bf{BHDG2:}} Remove the binary constraint on the hash matrix $B$, which makes $B$ numerical labels.
\end{itemize}
The results of these experiments are presented in Table 10. The key findings are summarized as follows: 

{\bf{Performance of BHDG1 vs. BHDG2}}: In most cases, BHDG2 outperforms BHDG1 across all evaluation metrics. This improvement can be attributed to the retention of the dynamic graph constraint component, which enhances feature selection by effectively constraining the sample space using the graph structure of the hashing matrix. 

{\bf{Performance of BHDG2 vs. BHDG}}: BHDG consistently outperforms BHDG2, suggesting that learning binary pseudo labels is crucial for achieving optimal performance. 

In summary, the ablation study confirms that each proposed component of BHDG is indispensable and contributes significantly to its overall effectiveness. The dynamic graph constraint and binary pseudo-label learning are particularly critical for enhancing classification.

\subsection{Analysis of convergence}
In this subsection, we examine the convergence behavior of the BHDG method across six datasets. As illustrated in Figure 19, the following observation can be made:

 {\bf{Rapid Convergence: For most datasets}}, the objective function exhibits a rapid decrease and stabilizes within 10 iterations. 
 
{\bf{Slower Convergence on Datasets}}: On the Reuters and Entertainment datasets, the algorithm requires up to 25 iterations to reach convergence. 

These results demonstrate that the proposed BHDG algorithm has strong convergence properties.

\section{Conclusion}
In this paper, we introduced a novel feature selection algorithm, Binary Hashing with Dynamic Graph Constraints (BHDG). This method integrates binary hashing learning and dynamic graph constraints to enhance feature selection performance.  Comprehensive experiments on 10 multi-label datasets demonstrate that BHDG consistently outperforms 10 competing methods, including RF-ML, MSSL, and MSFS. Ablation studies and convergence analysis further validate the contribution of each component of BHDG, highlighting its effectiveness and reliability.

The key advantages of BHDG are as follows: (1) Preserving Classification Structure: By converting numerical pseudo-labels into binary form, the algorithm effectively preserves the original classification structure, minimizing noise information. (2) Leveraging Manifold Structures: The dynamic graph constraints allow the model to explore the manifold structure of the sample projection space, facilitating the learning of more meaningful supervisory information. Despite these advantages, BHDG has certain limitations. Notably, recalculating the graph structure of the hashing matrix at each iteration introduces additional computational overhead. 

In future work, we aim to further optimize the efficiency of BHDG and address its computational limitations. Furthermore, we plan to investigate multi-label feature selection methods based on hyper-graph learning, which holds promise for large-scale data classification tasks.

\section*{Author Contributions}
Cong Guo: Writing-original draft, Conceptualization, Validation, Methodology; Changqin Huang: Writing-review and editing, Supervision. Wenhua Zhou: Writing-review and editing.  Xiaodi Huang: Writing-review and editing.
\section*{Conflicts of Interest}
The authors declare that they have no known competing financial interests or personal relationships that could have appeared to influence the work reported in this paper.
\section*{Availability of data and materials}
The datasets used in this study was obtained from the UCI and Mulan repositories and they are available in the following websites:\\ 1.http://archive.ics.uci.edu/ml/index.php\\
2.http://www.uco.es/kdis/mllresources/


\begin{thebibliography}{100}
\bibitem{A1}
P. Dhal and C. Azad, "A comprehensive survey on feature selection in the various fields of machine learning," Applied Intelligence, vol. 52, no. 4, pp. 4543-4581, 2022.
\bibitem{A2}
R. J. Urbanowicz, M. Meeker, W. La Cava, R. S. Olson, and J. H. Moore, "Relief-based feature selection: Introduction and review," Journal of biomedical informatics, vol. 85, pp. 189-203, 2018.
\bibitem{A3}
W. Qian, J. Huang, F. Xu, W. Shu, and W. Ding, "A survey on multi-label feature selection from perspectives of label fusion," Information Fusion, vol. 100, p. 101948, 2023.
\bibitem{A4}
J. Li et al., "Feature selection: A data perspective," ACM computing surveys (CSUR), vol. 50, no. 6, pp. 1-45, 2017.
\bibitem{A5}
C. Guo, W. Yang, C. Liu, and Z. Li, "Iterative missing value imputation based on feature importance," Knowledge and Information Systems, pp. 1-28, 2024.
\bibitem{A6}
C. Guo, W. Yang, Z. Li, and C. Liu, "A novel feature selection framework for incomplete data," Chemometrics and Intelligent Laboratory Systems, p. 105193, 2024.
\bibitem{A7}
Y. Zhang and Y. Ma, "Non-negative multi-label feature selection with dynamic graph constraints," Knowledge-Based Systems, vol. 238, p. 107924, 2022.
\bibitem{A8}
Y. Zhang, W. Huo, and J. Tang, "Multi-label feature selection via latent representation learning and dynamic graph constraints," Pattern Recognition, vol. 151, p. 110411, 2024.
\bibitem{A9}
J. Hu, Y. Li, G. Xu, and W. Gao, "Dynamic subspace dual-graph regularized multi-label feature selection," Neurocomputing, vol. 467, pp. 184-196, 2022.
\bibitem{A10}
L. Zhen, P. Hu, X. Wang, and D. Peng, "Deep supervised cross-modal retrieval," in Proceedings of the IEEE/CVF conference on computer vision and pattern recognition, 2019, pp. 10394-10403. 
\bibitem{A11}
L. Zhu, C. Zheng, W. Guan, J. Li, Y. Yang, and H. T. Shen, "Multi-modal hashing for efficient multimedia retrieval: A survey," IEEE Transactions on Knowledge and Data Engineering, vol. 36, no. 1, pp. 239-260, 2023.
\bibitem{A12}
Z. He, Y. Lin, Z. Lin, and C. Wang, "Multi-label feature selection via similarity constraints with non-negative matrix factorization," Knowledge-Based Systems, vol. 297, p. 111948, 2024.
\bibitem{A13}
J. Kileel, A. Moscovich, N. Zelesko, and A. Singer, "Manifold learning with arbitrary norms," Journal of Fourier Analysis and Applications, vol. 27, no. 5, p. 82, 2021.
\bibitem{A14}
L. Jian, J. Li, K. Shu, and H. Liu, "Multi-label informed feature selection," in IJCAI, 2016, vol. 16, pp. 1627-33. 
\bibitem{A15}
H. Li and H. Zhai, "Random Manifold Sampling and Joint Sparse Regularization for Multi-Label Feature Selection," Big Data Research, vol. 32, p. 100383, 2023.
\bibitem{A16}
Y. Li, L. Hu, and W. Gao, "Label correlations variation for robust multi-label feature selection," Information Sciences: An International Journal, 2022.
\bibitem{A17}
R. Shang, H. Chi, Y. Li, and L. Jiao, "Adaptive graph regularization and self-expression for noise-aware feature selection," Neurocomputing, vol. 535, pp. 107-122, 2023.
\bibitem{A18}
Z. Qin, H. Chen, Y. Mi, C. Luo, S.-J. Horng, and T. Li, "Multi-label Feature selection with adaptive graph learning and label information enhancement," Knowledge-Based Systems, vol. 285, p. 111363, 2024.
\bibitem{A19}
J. Ma, F. Xu, and X. Rong, "Discriminative multi-label feature selection with adaptive graph diffusion," Pattern Recognition, vol. 148, p. 110154, 2024.
\bibitem{A20}
Q. Zhou, Q. Wang, Q. Gao, M. Yang, and X. Gao, "Unsupervised Discriminative Feature Selection via Contrastive Graph Learning," IEEE Transactions on Image Processing, 2024.
\bibitem{A21}
 Y. Wang, X. Luo, and X.-S. Xu, "Label embedding online hashing for cross-modal retrieval," in Proceedings of the 28th ACM international conference on multimedia, 2020, pp. 871-879. 
\bibitem{A22}
D. Shi, L. Zhu, J. Li, Z. Zhang, and X. Chang, "Unsupervised adaptive feature selection with binary hashing," IEEE Transactions on Image Processing, vol. 32, pp. 838-853, 2023.
\bibitem{A23}
D. Lee and H. S. Seung, "Algorithms for non-negative matrix factorization," Advances in neural information processing systems, vol. 13, 2000.
\bibitem{A24}
Z. Lin, M. Chen, and Y. Ma, "The augmented lagrange multiplier method for exact recovery of corrupted low-rank matrices," arXiv preprint arXiv:1009.5055, 2010.
\bibitem{A25}
G. Tsoumakas, E. Spyromitros-Xioufis, J. Vilcek, and I. Vlahavas, "Mulan: A java library for multi-label learning," The Journal of Machine Learning Research, vol. 12, pp. 2411-2414, 2011.
\bibitem{A26}
A. Asuncion and D. Newman, "UCI machine learning repository," ed: Irvine, CA, USA, 2007.
\bibitem{A27}
N. Spolaôr, E. A. Cherman, M. C. Monard, and H. D. Lee, "ReliefF for multi-label feature selection," in 2013 Brazilian Conference on Intelligent Systems, 2013: IEEE, pp. 6-11. 
\bibitem{A28}
F. Nie, H. Huang, X. Cai, and C. Ding, "Efficient and robust feature selection via joint l2, 1-norms minimization," Advances in neural information processing systems, vol. 23, 2010.
\bibitem{A29}
J. Liu, S. Ji, and J. Ye, "Multi-task feature learning via efficient l2, 1-norm minimization," arXiv preprint arXiv:1205.2631, 2012.
\bibitem{A30}
 X. Chang, F. Nie, Y. Yang, and H. Huang, "A convex formulation for semi-supervised multi-label feature selection," in Proceedings of the AAAI conference on artificial intelligence, 2014, vol. 28, no. 1. 
\bibitem{A31}
Z. Cai and W. Zhu, "Multi-label feature selection via feature manifold learning and sparsity regularization," International journal of machine learning and cybernetics, vol. 9, pp. 1321-1334, 2018.
\bibitem{A32}
J. Zhang, Z. Luo, C. Li, C. Zhou, and S. Li, "Manifold regularized discriminative feature selection for multi-label learning," Pattern Recognition, vol. 95, pp. 136-150, 2019.
\bibitem{A33}
W. Gao, Y. Li, and L. Hu, "Multilabel feature selection with constrained latent structure shared term," IEEE Transactions on Neural Networks and Learning Systems, vol. 34, no. 3, pp. 1253-1262, 2021.
\bibitem{A34}
Y. Liu, H. Chen, T. Li, and W. Li, "A robust graph based multi-label feature selection considering feature-label dependency," Applied Intelligence, vol. 53, no. 1, pp. 837-863, 2023.
\bibitem{A35}
J. Hu, Y. Li, W. Gao, and P. Zhang, "Robust multi-label feature selection with dual-graph regularization," Knowledge-Based Systems, vol. 203, p. 106126, 2020.
\bibitem{A36}
M.-L. Zhang and Z.-H. Zhou, "ML-KNN: A lazy learning approach to multi-label learning," Pattern recognition, vol. 40, no. 7, pp. 2038-2048, 2007.
\bibitem{A37}
Z. Sun, H. Xie, J. Liu, and Y. Yu, "Multi-label feature selection via adaptive dual-graph optimization," Expert Systems with Applications, vol. 243, p. 122884, 2024.
\bibitem{A38}
J. Demšar, "Statistical comparisons of classifiers over multiple data sets," The Journal of Machine learning research, vol. 7, pp. 1-30, 2006.
\end{thebibliography}
\end{document}